\RequirePackage{amsthm}
\documentclass[iicol,sn-basic]{sn-jnl}

\usepackage{graphicx}%
\usepackage{multirow}%
\usepackage{amsmath,amssymb,amsfonts}%
\usepackage{amsthm}%
\usepackage[title]{appendix}%
\usepackage{xcolor}%
\usepackage{textcomp}%
\usepackage{manyfoot}%
\usepackage{booktabs}%
\usepackage{algorithm}%
\usepackage{algorithmicx}%
\usepackage{algpseudocode}%
\usepackage{listings}%
\usepackage[caption=false]{subfig}
\usepackage{booktabs}
\usepackage{flushend}
\usepackage{lmodern}
\usepackage{hyperref}
\usepackage{bookmark}

\theoremstyle{thmstyleone}%
\theoremstyle{thmstyletwo}%

\theoremstyle{thmstylethree}%

\begin{document}

\title[Article Title]{Temporal Correlation Meets Embedding: Towards a 2nd Generation of JDE-based Real-Time Multi-Object Tracking}

\author[1]{\fnm{Yunfei} \sur{Zhang}}\email{zhangyf32022@shanghaitech.edu.cn}
\equalcont{These authors contributed equally to this work.}
\author[2,3]{\fnm{Jin} \sur{Gao}}\email{jin.gao@nlpr.ia.ac.cn}
\equalcont{These authors contributed equally to this work.}
\author[4,7]{\fnm{Chao} \sur{Liang}}\email{chaoliang1996@gmail.com}
\equalcont{These authors contributed equally to this work.}
\author[5]{\fnm{Zhipeng} \sur{Zhang}}\email{zhipeng.zhang.cv@outlook.com}
\author[1,2,3]{\fnm{Weiming} \sur{Hu}}\email{wmhu@nlpr.ia.ac.cn}
\author[6]{\fnm{Stephen} \sur{Maybank}}\email{sjmaybank@dcs.bbk.ac.uk}
\author*[4,7]{\fnm{Xue} \sur{Zhou}}\email{zhouxue@uestc.edu.cn}
\author*[8]{\fnm{Liang} \sur{Li}}\email{liang.li.brain@aliyun.com}

\affil[1]{\orgdiv{School of Information Science and Technology}, \orgname{Shanghaitech University}, \orgaddress{\postcode{201210}, \state{Shanghai}, \country{China}}}
\affil[2]{\orgdiv{State Key Laboratory of Multimodal Artificial Intelligence System}, \orgname{Institute of
Automation, Chinese Academy of Sciences}, \orgaddress{\postcode{100190}, \state{Beijing}, \country{China}}}

\affil[3]{\orgdiv{School of Artificial Intelligence}, \orgname{ University of Chinese Academy of Sciences}, \orgaddress{\postcode{101408}, \state{Beijing}, \country{China}}}

\affil[4]{\orgdiv{Shenzhen Institute for Advanced Study}, \orgname{University of Electronic Science and Technology of China}, \orgaddress{\postcode{518110}, \state{Shenzhen}, \country{China}}}

\affil[5]{\orgdiv{KargoBot},\orgaddress{ \country{China}}}

\affil[6]{\orgdiv{Department of Computer Science and Information Systems}, \orgname{Birkbeck College}, \orgaddress{\state{WC1E 7HX London}, \country{U.K.}}}

\affil[7]{\orgdiv{School of Automation Engineering}, \orgname{ University of Electronic Science and Technology of China}, \orgaddress{\postcode{611731}, \state{Chengdu}, \country{China}}}

\affil[8]{\orgdiv{Beijing Institute of Basic Medical Sciences},  \orgaddress{\postcode{ 100850}, \state{Beijing}, \country{China}}}


\abstract{Joint Detection and Embedding (JDE) trackers have demonstrated excellent performance in Multi-Object Tracking (MOT) tasks by incorporating the extraction of appearance features as auxiliary tasks through embedding Re-Identification task (ReID) into the detector, achieving a balance between inference speed and tracking performance. However, solving the competition between the detector and the feature extractor has always been a challenge. Meanwhile, the issue of directly embedding the ReID task into MOT has remained unresolved. The lack of high discriminability in appearance features results in their limited utility. In this paper, a new learning approach using cross-correlation to capture temporal information of objects is proposed. The feature extraction network is no longer trained solely on appearance features from each frame but learns richer motion features by utilizing feature heatmaps from consecutive frames, which addresses the challenge of inter-class feature similarity. Furthermore, our learning approach is applied to a more lightweight feature extraction network, and treat the feature matching scores as strong cues rather than auxiliary cues, with an appropriate weight calculation to reflect the compatibility between our obtained features and the MOT task. Our tracker, named TCBTrack, achieves state-of-the-art performance on multiple public benchmarks, i.e., MOT17, MOT20, and DanceTrack datasets. Specifically, on the DanceTrack test set, we achieve 56.8 HOTA, 58.1 IDF1 and 92.5 MOTA, making it the best online tracker capable of achieving real-time performance. Comparative evaluations with other trackers prove that our tracker achieves the best balance between speed, robustness and accuracy. Code is available at \hyperlink{https://github.com/yfzhang1214/TCBTrack}{https://github.com/yfzhang1214/TCBTrack}.
}

\keywords{Multiple object tracking, cross-correlation, lightweight networks, re-identification}



\maketitle

\section{Introduction}\label{sec1}

Multi-Object Tracking (MOT) is a crucial task in computer vision, aiming to track multiple objects robustly and accurately over time. So far, MOT has been applied in various fields, including autonomous driving\citep{petrovskaya2009model,geiger2013vision,he2016precise,yu2020bdd100k,luo2021exploring,kim2022polarmot}, video surveillance \citep{milan2016mot16,dendorfer2020mot20}, and in high level tasks such as pose estimation and the prediction of human behaviour\citep{wu2019unsupervised,gu2019efficient}.

Tracking-by-Detection is one of the main approaches for addressing MOT tasks. It is a two-step process, with the detection of objects followed by the matching of objects\citep{kalman1960contributions,kuhn1955hungarian} between frames for object association. Within this paradigm, Joint Detection and Embedding (JDE) integrates the re-identification (ReID) task into the detector by adding a head branch to avoid re-extracting features from the ReID network. This integration has achieved a good balance between accuracy and speed \citep{wang2020towards,lu2020retinatrack,pang2021quasi,zhang2021fairmot,liang2022rethinking}. However, JDE trackers have an unresolved and significant shortage: directly using the same backbone as the detector results in small inter-class gaps in appearance features, which can lead to the lack of discrimination when similar objects appear in consecutive frames. 

\begin{figure}[t]
\centering
\begin{minipage}[t]{1.0\linewidth}
        \centering
        \includegraphics[width=2.8in]{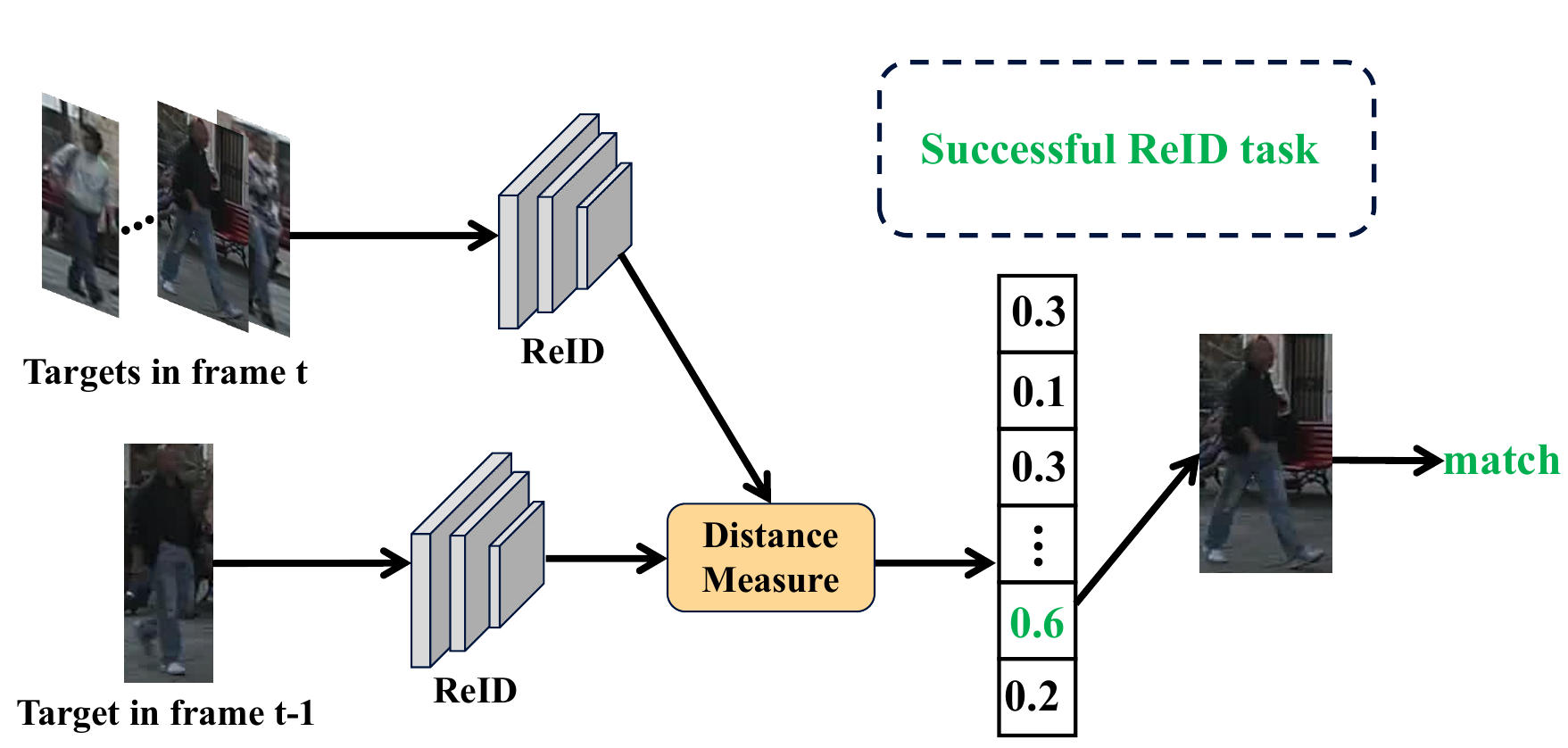}%
        \label{fig_sim1}
        \small \centerline{(a)}
        \end{minipage}%
\\
\begin{minipage}[t]{1.0\linewidth}
        \centering
        \includegraphics[width=2.8in]{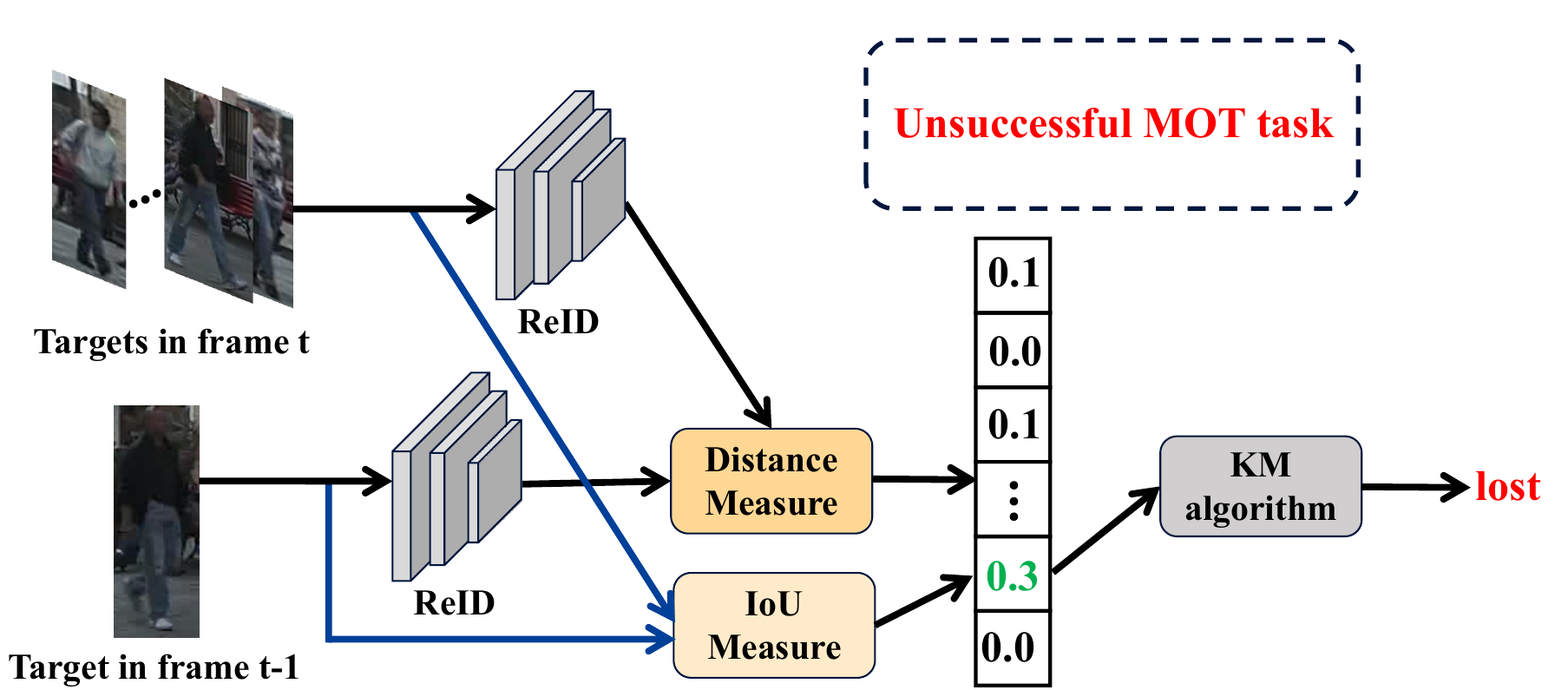}%
        \label{fig_sim2}
        \small\centerline{(b)}
        \end{minipage}%
\\
\caption{Comparisons between ReID task in MOT and MOT task. (a) ReID task in MOT. (b) MOT task. The ReID task can still be accomplished effectively when embedded into the MOT algorithm, while during the association stage, matching is not performed due to the small weights. This is caused by the training approach of the ReID model, leading to the loss of temporal information, and the misalignment of weights due to the blurry templates in MOT.}
\label{fig_sim}
\end{figure}

After revisiting the design process of all JDE trackers, we observe that one of the main reasons for small inter-class gaps in appearance features is that the high-dimensional features generated by the detector may not be suitable for feature extraction from specific objects. The detection backbone focuses more on the position information and classification of objects, while the feature extractor focuses more on the detailed features of objects. Previous works\citep{zhang2021fairmot,liang2022rethinking} balance these two tasks by designing richer feature extraction networks, which sacrifice speed in exchange for larger inter-class feature differences. Additionally, we have discovered similarities and differences between ReID and MOT. ReID and MOT uniquely identify individual objects. ReID finds all the images containing a given object. MOT associates target bounding boxes in the current frame with those from the previous frame. However, directly embedding the ReID task into the MOT task poses certain challenges, as illustrated in Fig. \ref{fig_sim}. From the figure, the ReID model may be successful within the MOT because it finds the bounding box most similar to the target. Nevertheless, when the MOT task employs linear assignments for matching, similar candidate boxes may not be matched due to their low scores in distance measure. This phenomenon is most prevalent when the template itself is blurry or when the retrieved bounding boxes are not accurate. Many works\citep{wang2020towards, zhang2021fairmot, liang2022rethinking,cui2023sportsmot} try to alleviate this issue by using linear weighting methods to obtain distance metrics and setting the weight of apparence features to a smaller value in order to assist in association. This approach leads to marginal performance improvements in tracking because the IoU metric is still an influential factor.


\begin{figure}[t]
\centering
\includegraphics[width=2.99in]{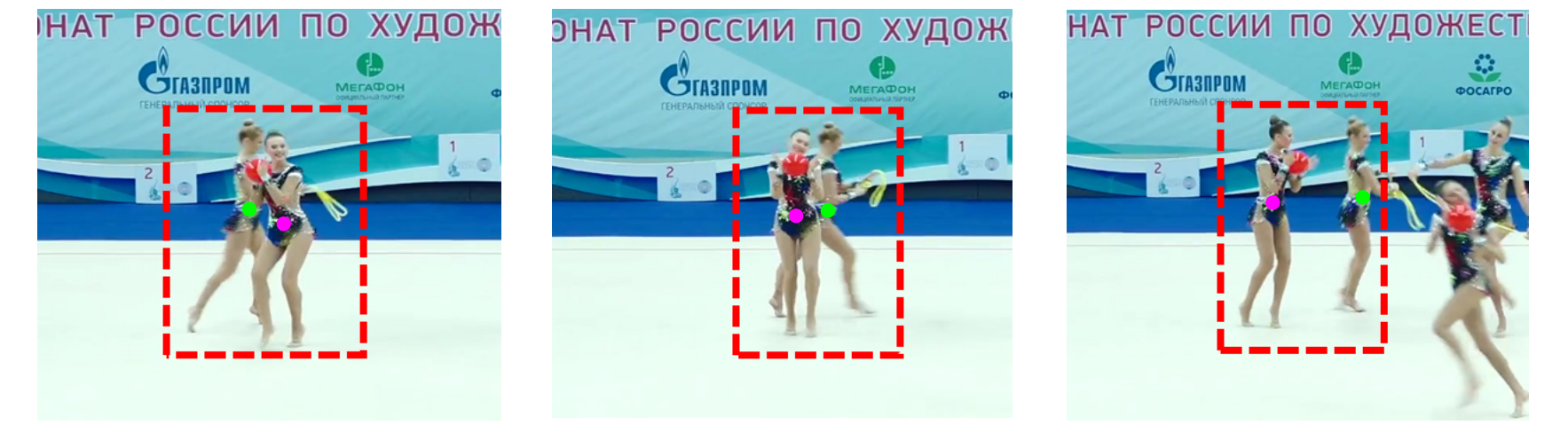}
\caption{Illustration of hard samples of JDE trackers. The dots in the dashed box represent the center of the object.}
\label{fig_hard}
\end{figure}

\begin{figure}[t]
\centering
\begin{minipage}[t]{1.0\linewidth}
        \centering
        \includegraphics[width=2.8in]{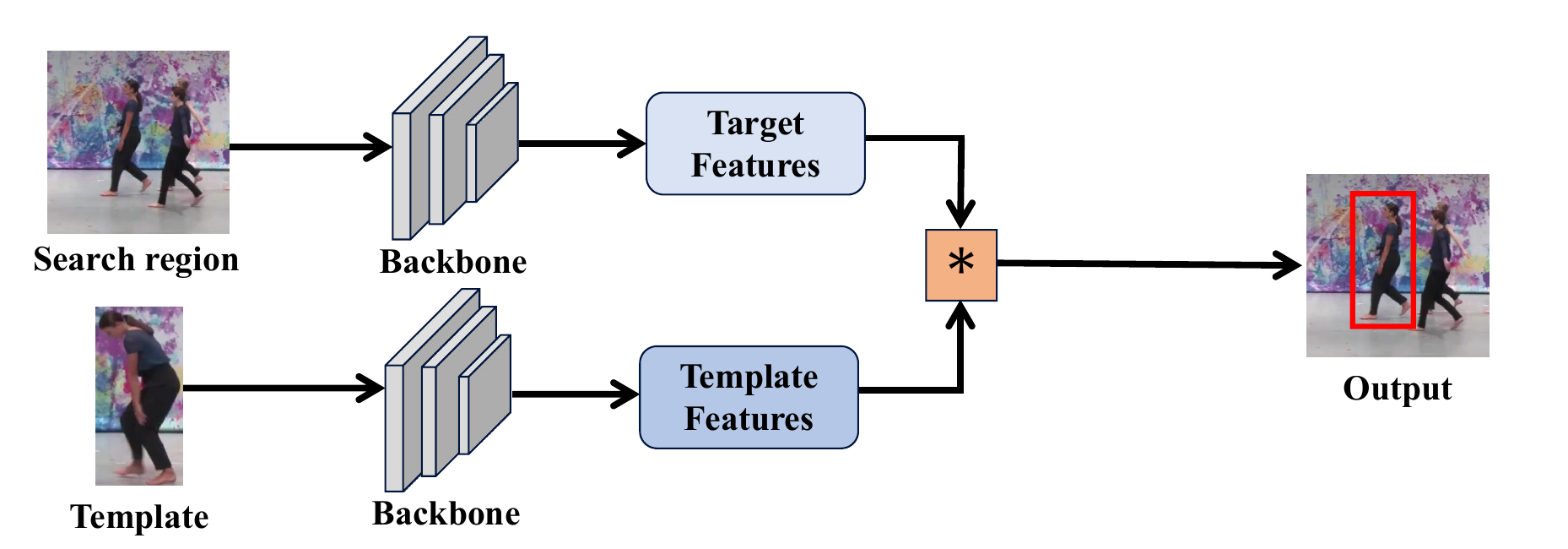}%
        \label{fig_1-1}
        \small \centerline{(a)Siamese network in SOT.}
        \end{minipage}%
\\
\begin{minipage}[t]{1.0\linewidth}
        \centering
        \includegraphics[width=2.8in]{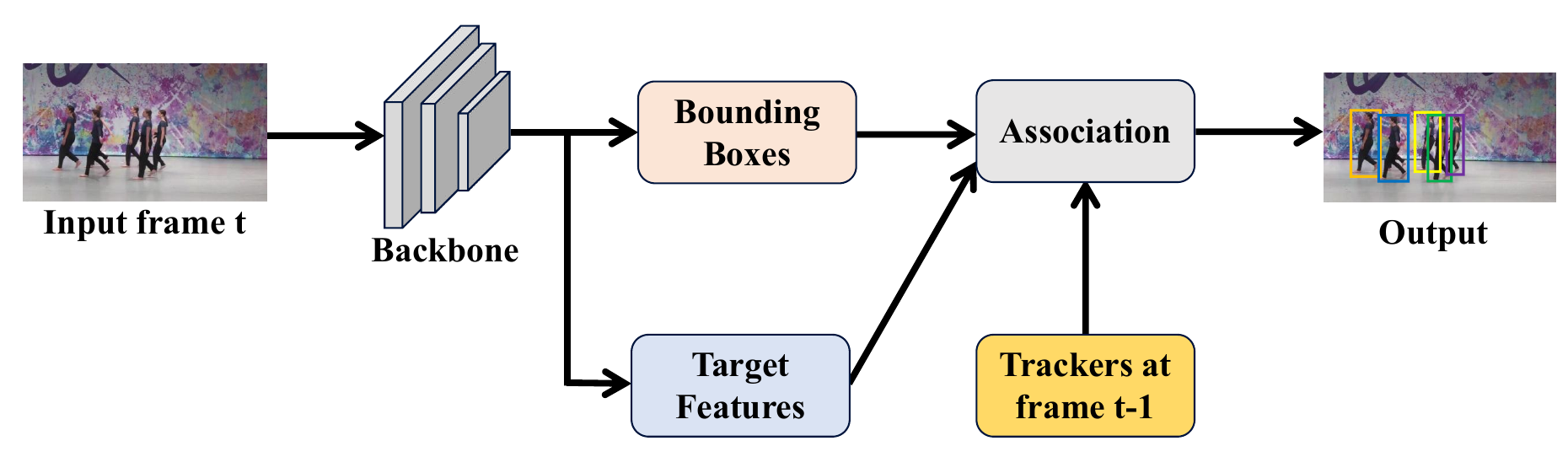}%
        \label{fig_1-2}
        \small\centerline{(b)JDE paradigm in MOT.}
        \end{minipage}%
\\
\begin{minipage}[t]{1.0\linewidth}
        \centering
        \includegraphics[width=2.8in]{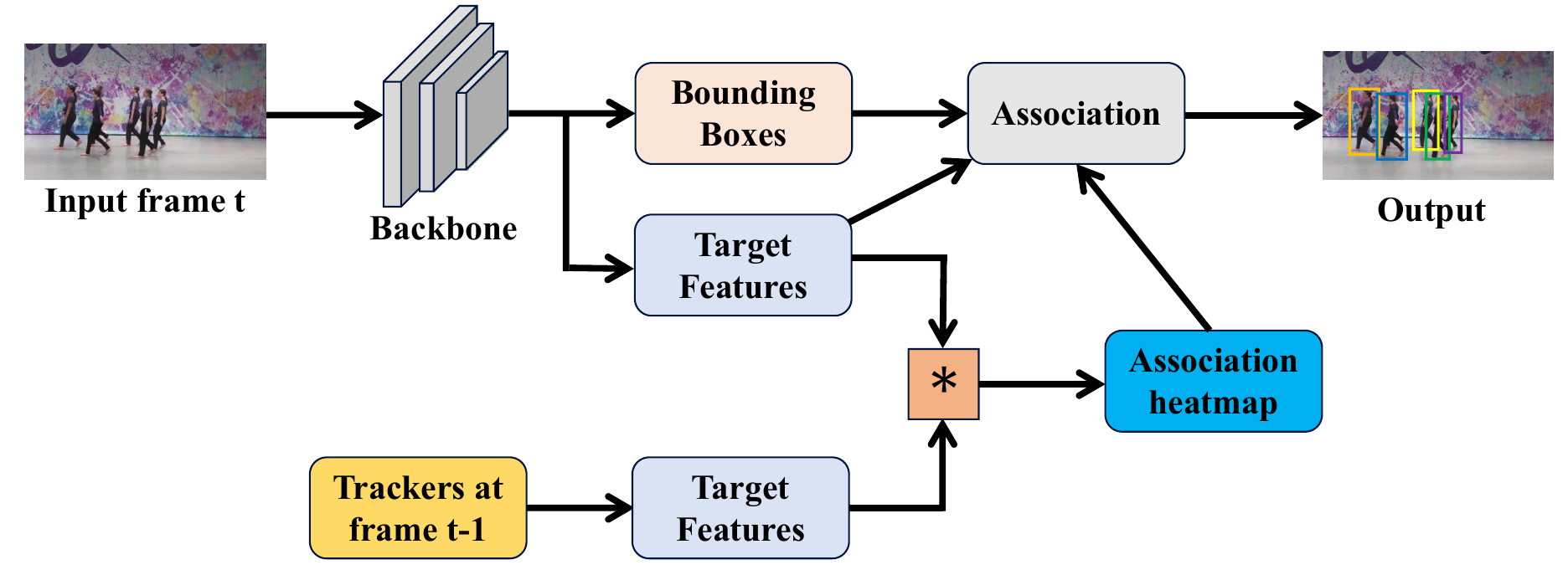}%
        \label{fig_1-3}
        \small\centerline{(c)Our method in MOT.}
        \end{minipage}%
\caption{The main framework of our MOT tracker. It involves incorporating object feature extraction inspired by SOT task into JDE structure, aiming to compensate for the limitations of feature extraction.}
\label{fig_1}
\end{figure}

We aim to develop a superior method for expressing appearance features. As shown in Fig. \ref{fig_hard}, when the appearance of objects is similar, it is difficult to compensate for the mismatch caused by the inaccurate IoU metric through the distance measure of apparence features. In fact, however, the object represented by the green dot in frames can be easily tracked, as the object has been moving to the right consistently. Many Single-Object Tracking (SOT) trackers\citep{varior2016siamese, li2018high,li2019siamrpn++} use template matching to establish motion feature responses between the template and the search region through cross-correlation operations, achieving localization of the tracked object. Inspired by siamese network in SOT, we incorporate cross-correlation into JDE structure, shown in Fig. \ref{fig_1}. Specifically, we construct frame training pairs through video training sets and use cross-correlation operations to obtain motion heatmap for each target. Gaussian function is used to generate corresponding groundtruth and apply Logistic-MSE loss. The learned temporal cues alleviate the problem of feature similarity between the object and its adjacent objects. At the same time, due to the motion information, the heatmap obtained by cross-correlation has a stronger expression in continuous frames. We treat the obtained heatmap as strong motion cues, and modify the distance measure in the association stage by replacing linear weighting with product, achieving better results. We have also significantly reduced the size of the feature extraction network to gain a speed advantage.

Finally, we have designed a more lightweight JDE-based tracker, named TCBTrack (Temporal Correlation-Based Tracker). By utilizing cross-correlation, we establish temporal relationships between the consecutive frames to enrich the appearance features. Furthermore, during the association stage, similarity scores and IoU scores are measured equally. As a result, we achieve improved tracking performance. Experiments are conducted on MOT17, MOT20, and DanceTrack datasets\citep{milan2016mot16,dendorfer2020mot20,sun2022dancetrack}, and achieve state-of-the-art results. Compared to other trackers, our tracker still exhibits the best balance in terms of robustness, accuracy, and speed.

This paper is an extension of our conference version published in AAAI'22\citep{liang2022one}. Our Previous work, OMC tracker, introduces a temporal cue training module and a re-check module. OMC applies them to recall low-confidence detection boxes to compensate for the shortcomings in detection performance. We believe that there are opportunities for research in this area. We make the following improvements to \cite{liang2022one}: 
\begin{itemize}
    \item The object detector and the appearance model are improved. With the advancement of detectors, occluded and blurred objects can be detected more reliably, which undoubtedly provides the greatest assistance to Tracking-by-Detection paradigm trackers. Richer image features also have a promoting effect on the appearance model, so we further lightweight the appearance model.
    \item Temporal information from OMC is included. We have removed the re-check module used for recalling detection boxes in OMC, as the original issue has already been addressed by existing good detectors.
    \item A better training method for extracting the temporal features of objects is implemented, alleviating the feature similarity issue among similar objects and addressing the convergence difficulties of the ReID task in large training datasets. This approach is then applied to the appearance model.
    \item The association of object features over time is improved. An improved calculation method for matching scores is set up.
    \item The new tracker is tested on the difficult DanceTrack dataset\citep{sun2022dancetrack} and more ablation studies are added. 
    \item More comparisons with existing trackers are described.
\end{itemize}

\section{Related Work}\label{sec2}
\subsection{Tracking-by-Detection Paradigm}
Over the years, Tracking-by-Detection paradigm has been the focal point of MOT tasks. After object detection it is necessary to associate objects in different frames. Early MOT methods relied on the best available detectors at the time to enhance association accuracy. IOU tracker\citep{bochinski2017high} achieves fast multi-object tracking by matching objects in adjacent frames based on Intersection over Union (IoU)\citep{yu2016unitbox} as the matching criterion. SORT\citep{bewley2016simple}, which is an improvement on the IoU tracker, introduces Kalman filter\citep{kalman1960contributions} and Hungarian matching algorithm\citep{kuhn1955hungarian} to enhance the accuracy of the IoU-based association algorithm by predicting the objects in the next frame and reducing ID switches. DeepSORT\citep{wojke2017simple} further enhances SORT by adding a feature extraction module, which extracts ReID features for each object by feeding the detected bounding boxes into a deep neural network. It then excludes associations of objects with significant distances between features, as measured by the Mahalanobis distance\citep{mahalanobis2018generalized}. It is noted that early tracking tasks assume simple datasets, where the combination of Kalman Filter and IoU metric is sufficient to solve most association matching problems. Therefore, many works use ReID features as a weak cues to help eliminate matches with similar positions and sizes of bounding boxes but distinct appearance features. See \cite{yu2016poi}, \cite{wojke2017simple}, \cite{tang2017multiple}, \cite{chen2018real}, \cite{aharon2022bot}, \cite{wang2022smiletrack}, \cite{du2023strongsort}, \cite{cui2023sportsmot}, \cite{yang2023hybrid}. With the improvement of ReID techniques, differences in appearance features among different objects have become more prominent. Many trackers have utilized appearance information to obtain higher-level information and enhance object association. See \cite{zhang2019theoretically}, \cite{zhou2020tracking}, \cite{wang2021multiple}, \cite{dai2021learning}, \cite{chu2023transmot}. However, as trackers strive for higher accuracy, they introduce significant additional computational overheads, making it challenging to balance precision and speed. JDE\citep{wang2020towards} proposes the integration of feature extraction as a detection branch, reducing computational requirements by sharing the backbone with detectors and achieving good results in terms of performance and speed. FairMOT\citep{zhang2021fairmot} finds JDE-based trackers which utilize a detection-learned backbone, thereby addressing the fairness issue in feature extraction and establishing a new feature extraction network to balance the detection and feature extraction tasks. Many other trackers have similar strategies. See  \cite{bergmann2019tracking},\cite{lu2020retinatrack}, \cite{pang2021quasi}, \cite{liang2022rethinking}. From these developments, it is evident that extracting features efficiently and accurately has become the crucial objective of Tracking-by-Detection.

\subsection{MOT Methods With Feature Extraction}
For methods that rely solely on bounding boxes for motion prediction\citep{bewley2016simple,zhang2022bytetrack,yi2023ucmctrack,cao2023observation,yang2023hybrid,liu2023sparsetrack}, the quality of the detection boxes is crucial. However, there may be errors in the detection boxes if there are high speed object motions or large occlusions. To address this degradation, some approaches incorporate temporal cues from previous and future frames. For instance, \cite{braso2020learning}, \cite{cetintas2023unifying} leverage graph neural networks to establish object connections between frames. However, these methods require the entire video sequence as input to capture global temporal associations, which makes them unsuitable for online tracking. 

Other methods\citep{zhang2021fairmot,pang2021quasi,yang2021remot,liang2022rethinking,aharon2022bot,du2023strongsort,maggiolino2023deep} enhance deep learning networks to obtain appearance features by employing improved feature extraction networks to increase inter-class differences. Most JDE Trackers use a multi-layer perceptron (MLP) to map to the ID dimensions during training and employ ID loss for learning \citep{wang2020towards,zhang2021fairmot,liang2022rethinking,yu2022relationtrack}. As a consequence, as the training sample size increases, the structure of the MLP also changes due to the increase of objects, making it difficult for the feature extractor to converge when dealing with the large training sets. Additionally, the feature extraction network remains limited to single frames in which ReID only learns appearance. As a result, features generated through the feature extraction network are not effective in discerning multiple overlapped or closely positioned objects and only play an auxiliary role during the association stage. Some methods\citep{aharon2022bot,du2023strongsort,maggiolino2023deep} also use triplet loss\citep{schroff2015facenet} to train the ReID model. Nevertheless, selecting effective triplets can be challenging for JDE trackers in MOT datasets, particularly in the presence of occlusions, motion crossovers, and scene variations: the low sensitivity of triplet loss to hard samples may lead to instability and degraded performance of the feature extractor(See \cite{wang2020towards}).


Additionally, there are works that integrate SOT methods into the MOT task. This is because SOT trackers exhibits higher requirements for feature details during matching and faces common challenges in tracking tasks, such as occlusions and irregular motion. Among them, trackers that use siamese network utilize feature matching to locate the object position, which it is not limited by the detection. Early trackers\citep{chu2017online,sadeghian2017tracking,zhu2018online,zheng2021improving} draw inspiration from SiamFC\citep{bertinetto2016fully} and employ siamese networks to generate temporal information, replacing the traditional appearance modeling algorithms for MOT associations. Unfortunately, these trackers have a high computational complexity, because each located object has to be compared with all the tracked objects. This limitation makes them unsuitable for real-time applications. Also inspired by SOT, our previous work OMC\citep{liang2022one} is the first to use a siamese network and similarity matching to obtain reliable information from poorly detected bounding boxes. OMC incorporates object detection, temporal feature extraction, and object recall into a unified network, making it suitable for online real-time applications and transferrable to other trackers of the same Tracking-by-Detection paradigm to improve performance. With recent advancements in object detection models\citep{girshick2015fast,ren2015faster,he2017mask,lin2017feature,lin2017focal,redmon2018yolov3,ge2021yolox}, the performance boost obtained by including poorly detected bounding boxes is reduced. To explore other expressions of the siamese network, we introduce a refined version of OMC, named TCBTrack, which replaces the object detection backbone with a new, more efficient one and uses similarity matching for training and association. Unlike other traditional ReID learning methods, similarity matching not only learns appearance information but also temporal information to make the features more discriminative. By emphasizing appearance score in the association algorithm, the performance of the tracker is further enhanced. In summary, our approach utilizes siamese network architectures to extract temporal features, enabling deployment with both subpar and strong detectors while maintaining robustness and high-speed capability. 

\subsection{Other MOT Methods}
Vision transformer (ViT)\citep{dosovitskiy2020image} is an example of a transformer that has been applied to many computer vision tasks\citep{carion2020end,hatamizadeh2022unetr,zhai2022scaling}. It introduces the concept of treating image patches as tokens and extracting image features using transformers, demonstrating the capability of transformers in the domain of image extraction. The Detection Transformer (DETR)\citep{carion2020end} further expands the applicability of transformers in computer vision. It considers the input to the encoder part of the transformer as object queries, which are obtained by summing learnable positional encodings with image features extracted through a CNN network. The decoder part, along with prediction heads, then generates bounding boxes and confidence scores for each query, enabling end-to-end object detection. Subsequently, the Multiple Object Tracking Transformer (MOTR)\citep{zeng2022motr} introduces modifications to the DETR model which are specifically tailored for MOT tasks. Traditional detection-based tracking algorithms often separate the detection and association tasks. The associations of the objects often rely on traditional mathematical methods that lack generality. MOTR proposes leveraging a similar model structure to DETR to incorporate temporal modeling capabilities for MOT tasks.

MOTR is a popular framework for end-to-end object tracking. MOTRv2\citep{zhang2023motrv2}, for example, extends MOTR by combining it with YOLOX\citep{ge2021yolox}, resulting in an improvement in tracking performance. Besides, there are other similar methods such as \cite{li2023end}, \cite{yu2023motrv3},  \cite{gao2023memotr} which have also demonstrated the success of the DETR paradigm in object tracking tasks.

Based on the ViT architecture, tracking frameworks show promising performance. Nevertheless, there are still difficulties, for example the stacking of the transformer networks and the requirement of long sequences of features as token inputs. End-to-End models consistently suffer from a speed disadvantage\citep{liu2023ring,liu2024blockwise}. This is because the object association between frames and object detection are learned within the same network, necessitating more complex and larger models in support. To address this issue, some works propose using Graph Neural Networks (GNNs) for inter-frame object association, such as \cite{braso2020learning}, \cite{liu2020gsm}, \cite{li2020graph}, \cite{papakis2020gcnnmatch}, \cite{cetintas2023unifying}. These approaches employ GNNs to compute object box information (e.g., aspect ratio, size, position, confidence) and derive inter-frame weights or establish trackers between consecutive frames, enabling concatenation through the GNN framework. Unfortunately, these GNN-based methods require the entire video sequence as input, rendering them incapable of achieving online tracking. Currently, researchers are actively exploring more efficient approaches to tackle online object tracking. All types of trackers have been classified as shown in Table \ref{table:1}.

\begin{table}[!t]
\caption{MOT method comparison.}\label{table:1}%
\begin{tabular}{@{}llll@{}}
\toprule
\textbf{Algorithms} & \textbf{speed} & \textbf{online} & \textbf{robust} \\  
\midrule
Transformer-based & $\times$ & \checkmark & \checkmark  \\ 
TBD-based\footnotemark[1](motion) & \checkmark & \checkmark & $\times$  \\
TBD-based\footnotemark[1](SDE\footnotemark[2]) & $\times$ & \checkmark & \checkmark  \\      
TBD-based\footnotemark[1](JDE) & \checkmark & \checkmark & \checkmark  \\
GNN-based & \checkmark & $\times$ & \checkmark  \\ 
\botrule
\end{tabular}
\footnotetext[1]{TBD: Abbreviation of ``Tracking-by-Detection''}
\footnotetext[2]{SDE: Abbreviation of ``Separate Detection and Embedding''}
\end{table}

\section{Methodology}
\begin{figure*}[!t]
\centering
\includegraphics[width=6.2in]{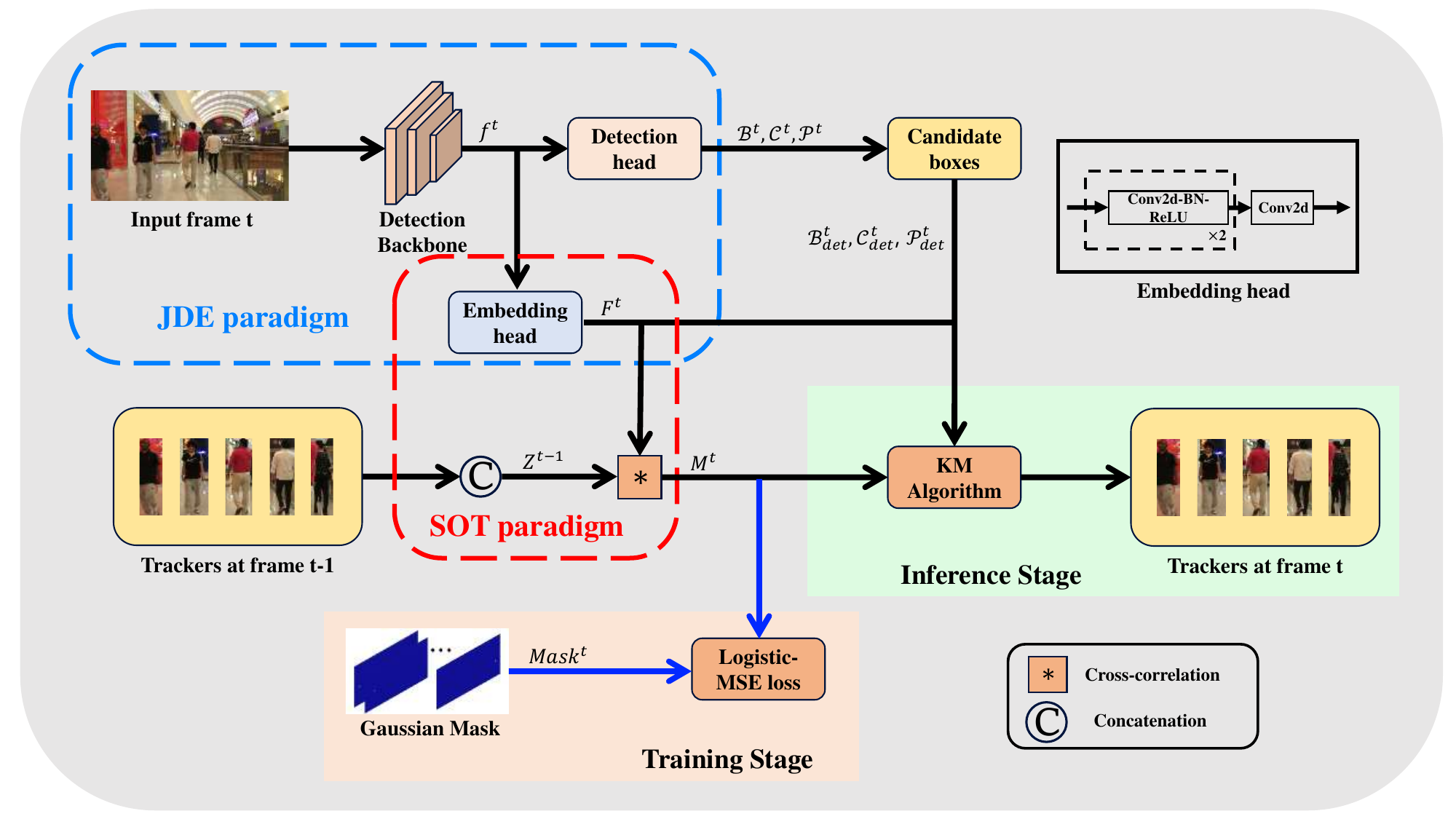}
\caption{Overview of the proposed TCBTrack. It consists of the JDE paradigm, the SOT paradigm and our training/inference stage(the blue arrows represent the process during the training stage.). The Detection Backbone first generates the detection result $\mathcal{B}^t, \mathcal{C}^t, \mathcal{P}^t $and candidate embeddings $F^{id}_t$. We use a Gaussian heatmap and Logistic-MSE loss in the training stage, and the joint scores of heatmap, IoU, and confidence are used as weights in the inference stage. }
\label{fig_omcv1}
\end{figure*}
\bmhead{Tracking-by-Detection}In the inference stage, one single video frame $I^{t}$ is fed into the neural network. Similar to conventional Tracking-by-Detection methods, a high-dimensional feature map $f^{t}$ is extracted from $I^{t}$ using the detection backbone $\Phi$. Subsequently, object detection and prediction of bounding boxes and confidence scores are achieved by the detection head $\Psi$ represented as:
\begin{equation}
\label{1}
    f^{t} = \Phi(I^{t}),
\end{equation}
\begin{equation}
\label{2}
    [\mathcal{B}^t, \mathcal{C}^t, \mathcal{P}^t] = \Psi(f^{t}).
\end{equation}
$\mathcal{B}^t$$\in$$\mathbb{R}^{H\times W \times 4}$ is the predicted position of the objects after the image frame passes through the backbone and head, $\mathcal{C}^t$$\in$$\mathbb{R}^{H\times W \times class}$ is the class probability prediction for each bounding box, and $\mathcal{P}^t$$\in$$\mathbb{R}^{H\times W \times 1}$ is the confidence prediction for each bounding box. To efficiently obtain feature vectors for each independent candidate box, we adopt the feature extraction method from JDE, which involves sharing the backbone required for feature extraction with the detector, and incorporating an additional head branch $\Omega$ for feature extraction as follows:
\begin{equation}
\label{3}
    F^{t} = \Omega(f^{t}),
\end{equation}
where $F^t$$\in$$\mathbb{R}^{H\times W \times C_{out}}$ represents the feature vectors corresponding to all candidate boxes, with each bounding box $F^{t}_{i}$$\in$$\mathbb{R}^{1\times 1 \times C_{out}}$ possessing a feature of dimension $C_{out}$.

After obtaining the candidate box positions, confidence scores, classification predictions, and feature vectors for $I^{t}$, all candidate boxes are subjected to non-maximum suppression (NMS) to filter out the final candidate boxes based on a combination of confidence and classification probability, expressed as:
\begin{equation}
\label{4}
    \mathcal{S}^{t}_{i} = \mathcal{P}^{t}_{i} \cdot max(\mathcal{C}^{t}_{i}(0),\mathcal{C}^{t}_{i}(1),...,\mathcal{C}^{t}_{i}(class-1))|_{i=1}^{n},
\end{equation}
\begin{equation}
\label{5}
    [\mathcal{B}^t_{det}, \mathcal{C}^t_{det}, \mathcal{P}^t_{det}] = NMS([\mathcal{B}^t, \mathcal{C}^t, \mathcal{P}^t], \mathcal{S}^{t})
\end{equation}

$\mathcal{S}^{t}_{i}$ is the overall score of each bounding box. The filtered candidate boxes are used in the tracking association stage. We have defined an embedded feature extraction method that performs well for any detector; this ensures that we can easily achieve a lightweight tracker to meet real-time requirements while ensuring accuracy. Furthermore, we simplify the embedding head with our new learning method, enabling the model to utilize better detectors under speed requirements.

\subsection{TCBTrack architecture}

The accuracy of the detector is a significant factor in Track-by-Detection. When the accuracy of the detector is low, the bottleneck in MOT is the setting of a threshold for detecting bounding boxes. When the accuracy of the detector is high, the bottleneck in MOT is the description of the complex motions of objects. It is traditional to use the Kalman filter to predict the positions of objects in the next frame\citep{bewley2016simple,yu2016poi,bochinski2017high,bochinski2018extending,wang2020towards,zhang2022bytetrack}. However, Kalman filtering is not robust in complex environments. Our inspiration comes from the experimental results of the low-detection model\citep{liang2022one}: the heatmap obtained through cross-correlation learning can be used to find objects missed by the detector. Embedding the appearance model into MOT does not provide significant improvements, hence most methods still use IoU as the core because it is considered as strong cues for matching. Cross-correlation, aiming to utilize the motion information and appearance information of the object for feature matching, can also provide strong cues for association. Therefore, we also have enhanced the proportion of our generated features obtained in the association stage to achieve better tracking performance. The entire training stage and inference stage of TCBTrack can be seen in Fig. \ref{fig_omcv1}.

\bmhead{Embedding head} In JDE, the embedding head transforms the high-dimensional features used for detection into unique feature vectors for the objects. Just as the detector outputs the confidence, classification, and detection box position of the objects through multiple head branches, the feature vectors of the candidate boxes can also be obtained after the detection backbone. Considering the importance of the detector for MOT, we choose to simplify the computational overhead of the embedding head as much as possible and obtain better performance through specific training methods. Here, we choose to set the embedding head composed of two Conv2d-BN-ReLU layers with 3*3 convolutional kernels, and use one 1*1 convolutional kernel Conv2d to map the dimensions to the dimension of the feature vector(we take dim=512).

\bmhead{Transductive Detection Module} The transductive detection module aggregates the object features from the previous frame and the candidate box features from the current frame to obtain a temporal cue map. As in SiamFC there are \textit{k} templates, denoted as $z^{t-1}_{1}$, $z^{t-1}_{2}$, ..., $z^{t-1}_{k}$. After operating equation \eqref{3}, we obtain the feature vectors $F^{t}$ for all candidate boxes in the current frame. Therefore, we use the operator * for box localization:
\begin{equation}
\label{6}
    m^{t}_{i} = z^{t-1}_{i}* F^{t}|_{i=1}^{n}.
\end{equation}
$m^{t}_{i}$$\in$$\mathbb{R}^{H \times W}$ represents the heatmap of the i-th tracker at frame \textit{t}. Through \textit{k} operations and concatenation, we can obtain the object localization information $\mathcal{M}^{t}=\{m^{t}_{1}, m^{t}_{2}, ..., m^{t}_{k}\}$ for all templates in the current frame. It is worth noting that, since the dimension of the template is $\mathbb{R}^{1\times 1 \times C_{out}}$, the cross-correlation operation can be modified to a simple matrix multiplication. Therefore, we can concatenate the features of all trackers to save time and obtain the heatmap for each template. The implementation details of the cross-correlation operation can be found in Algorithm \ref{ag1}.
\begin{algorithm}[H]
\caption{Pseudocode of cross-correlation, Pytorch.}
\begin{algorithmic}[1]
\State {\textbf{def} generate\_heatmap}$(Z^{t-1}, F^{t}):$
\State  Z = $Z^{t-1}$.view(k, cout)
\State  Z\_norm = torch.linalg.norm(Z, dim=1, keepdim=True) 
\State  F = $F^{t}$.view(H*W, cout)
\State  F\_norm = torch.linalg.norm(F, dim=1, keepdim=True) 
\State  M = torch.matmul(Z, F.T) 
\State  M = torch.div(M, torch.matmul(Z\_norm, 
\Statex \hspace{0.7cm} F\_norm.T)
\State  \textbf{return} M.view(cout, H, W) 
\end{algorithmic}
\label{ag1}
\end{algorithm}

\bmhead{Training} We do not make any changes in the detection model so the backbone and head of the detector are frozen throughout the training process. To train our embedding head, the following steps are followed:

1)Image pair selection. We randomly select image pairs such that the two images in each frame are separated by no more than five frames, which generates more image pairs for model training and ensures that the majority of the objects in the two images can be matched. Additionally, choosing images frames at slightly larger intervals(e.g., interval with 5 frames) allows the model to learn richer temporal information, as the intervals reduce reliance on appearance information. This helps the model to generate high response heatmaps even when the object is moving rapidly or undergoing deformation.

2)Object selection. Firstly we perform a forward process on the t-1 frame's image $I^{t-1}$ to obtain all candidate boxes $\mathcal{B}^{t-1}$, $\mathcal{C}^{t-1}$, $\mathcal{P}^{t-1}$, and their corresponding feature vectors $F^{t-1}$ from the previous frame(equation \eqref{2}, equation \eqref{3}). Subsequently, we use IoU 
 in combination with ground truth $\mathcal{B}^{t-1}_{gt}$ and $\mathcal{ID}^{t-1}_{gt}$ for candidate box selection, thereby obtaining object box features and IDs of the previous frame, as detailed in Algorithm \ref{ag2}).

\begin{algorithm}[!ht]
\caption{Pseudocode of object selection.}
\begin{algorithmic}
    \State \hspace{0.2cm}{\textbf{Input}}: All bounding boxes of the previous frame $\mathcal{B}^{t-1}$, ground truth bounding boxes $\mathcal{B}^{t-1}_{gt}$, features of the bounding boxes $F^{t-1}$, ground truth IDs $\mathcal{ID}^{t-1}$, threshold $\alpha$.
    \State \hspace{0.2cm}{\textbf{Output}}: Features of the selected objects $F^{t-1}_{sel}$, IDs of the selected objects $\mathcal{ID}^{t-1}_{sel}$.\vspace{0.1cm}

\end{algorithmic}

\begin{algorithmic}[1]
    \State  $index_{1}$ = argmax(IoU($\mathcal{B}^{t-1}$,$\mathcal{B}^{t-1}_{gt}(i)$))$|_{i=1}^{n}$
    \State  scores = max(IoU($\mathcal{B}^{t-1}$,$\mathcal{B}^{t-1}_{gt}(i)$))$|_{i=1}^{n}$
    \State  $index_{2}$ = scores$>\alpha$
    \State  $\mathcal{ID}^{t-1}_{sel}$ $\gets$ $index_{2}$
    \State   $F^{t-1}_{sel}$ $\gets$ [$index_{1}$,$index_{2}$]
    \State \Return [$F^{t-1}_{sel}$,$\mathcal{ID}^{t-1}_{sel}$]
\end{algorithmic}
\label{ag2}
\end{algorithm}

The process first utilizes each $b^{t-1}_{gt}\in \mathcal{B}^{t-1}_{gt}$ to generate IoU response on $\mathcal{B}^{t-1}$, then selects the candidate box with the highest response value and its high-dimensional feature, labeling it with the ID number. If the IoU response is less than $\alpha$(set to 0.8), it signifies that the detector itself cannot produce correct candidate boxes, and such objects will not be considered. When the score is greater than $\alpha$, it indicates that the detector has a bounding box capable of capturing the object (regardless of its classification probability and confidence), and therefore can be used for training. By performing the similar operation, we obtain all candidate boxes $\mathcal{ID}^{t}_{sel}$ for the next frame $I^{t}$ along with their corresponding $\mathcal{B}^{t}$, $\mathcal{B}^{t}_{gt}$, $\mathcal{ID}^{t}_{gt}$. Candidate boxes with the same ID across two frames are considered as training pairs:
\begin{equation}
\label{13}
    Index = \{i|\mathcal{ID}^{t-1}_{sel}(i)\in \mathcal{ID}^{t}_{sel}\},
\end{equation}
\begin{equation}
\label{14}
    [\Tilde{F}^{t-1}_{sel}, \Tilde{\mathcal{ID}}^{t-1}_{sel}] \gets [F^{t-1}_{sel},\mathcal{ID}^{t-1}_{sel},Index].
\end{equation}

Additionally, through the information of IDs $\Tilde{\mathcal{ID}}^{t-1}_{sel}$, the groundtruth will be updated to $\Tilde{\mathcal{B}}^{t}_{gt}$, $\Tilde{F}^{t}_{gt}$, $\Tilde{\mathcal{ID}}^{t}_{gt}$ applicable to the training process.

3)Model forward and groundtruth generation. We perform cross-correlation computation between the obtained $F^{t}$ and $\Tilde{F}^{t-1}_{sel}$ calculated from the current frame to obtain the heatmap for each object using equation \eqref{6}. We use Gaussian distributions to generate the label for each groundtruth, and by concatenation, we obtain the groundtruth heatmap $H^t$. A single groundtruth heatmap can be expressed by the following formula:
\begin{equation}
\label{15}
    \resizebox{.8\hsize}{!}{$h^t_{i}\in \mathbb{R}^{H\times W} = exp(-\frac{(x-cx_i)^2+(y-cy_i)^2}{2\sigma^2})$}
\end{equation}
where the location ($cx_i$, $cy_i$) can be obtained from $\Tilde{\mathcal{B}}^{t}_{gt}(i)$, and $\sigma$ is the object size-adaptive standard deviation. We do not assign the central point as 1 and assign the region outside the central point as 0 . This is because adjacent candidate boxes that represent the same object contain slightly similar temporal information and cannot be rudely considered as negative samples.

\begin{figure*}[!ht]
        \centering
        \begin{minipage}[t]{0.33\linewidth}
        \centering
        \includegraphics[width=2.0in]{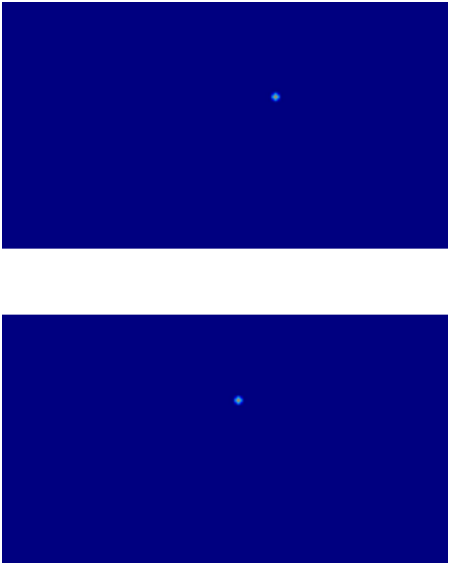}%
        \label{fig_loss1}
        \small\centerline{(a)}
        \end{minipage}%
        \begin{minipage}[t]{0.33\linewidth}
        \centering
        \includegraphics[width=2.0in]{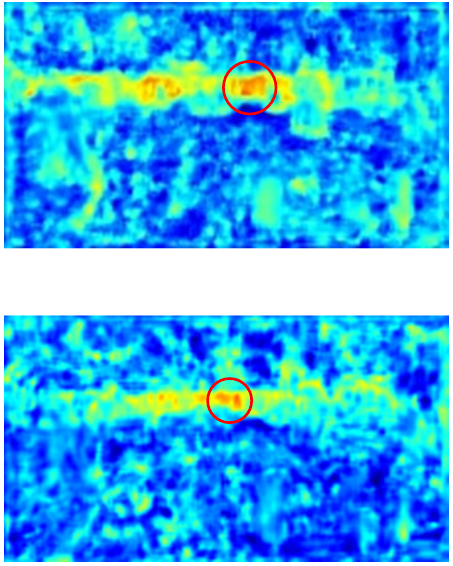}%
        \label{fig_loss2}
        \small\centerline{(b)}
        \end{minipage}%
        \begin{minipage}[t]{0.33\linewidth}
        \centering
        \includegraphics[width=2.0in]{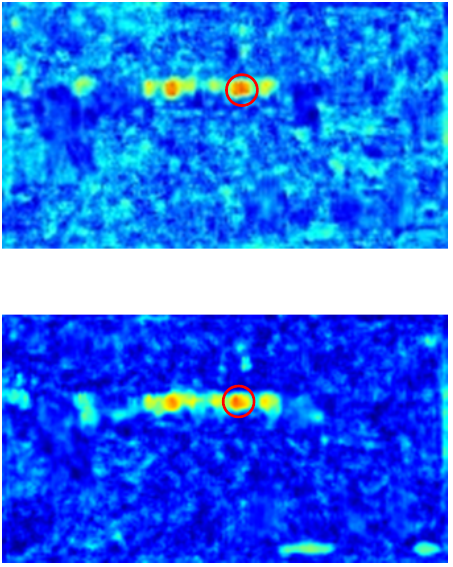}%
        \label{fig_loss3}
        \small\centerline{(c)}
        \end{minipage}%
        \centering
        \caption{Sample heatmap results using ID loss and our Logistic-MSE loss. (a) Groundtruth heatmap. (b) Heatmap generated by ID loss. (c) Heatmap generated by Logistic-MSE loss. The area circled in red represents the highest response of the heatmap. Our method can suppress the surrounding response of the target box and alleviate the problem of incorrect matching between adjacent objects.}
        \label{fig_loss123}
\end{figure*}

4)Loss design. The calculation is similar to the siamese method for calculating the loss on the heatmap. By using cross-entropy loss between the heatmaps generated by the model and the groundtruth heatmap generated through Gaussian distribution, we can suppress the localization of objects in other positions, thereby learning temporal cues. We use Logistic-MSE Loss as our loss function, and the specific formula is as follows: 
\begin{equation}
\label{19}
    \resizebox{.95\hsize}{!}{$\mathcal{L}_{temp} = -\frac{1}{n}\sum_{(m,h)}\sum_{x,y} \left\{
\begin{aligned}
& (1-m_{x,y})log(m_{x,y}), \text{if }h^t_{x,y}\geq 1\\
& (1-h^{t}_{x,y})m_{x,y}log(1-m_{x,y}) , \text{else}. \\
\end{aligned}
\right.$}
\end{equation}
$m_{x,y}$ represents every single point of every heatmap. We do not add any other losses to demonstrate the benefits of our training method, so the final loss can be simply represented as $\mathcal{L}_{temp}$. Visualization results are shown in Fig. \ref{fig_loss123}. Although the surrounding candidate boxes have been suppressed, there are still other high response values. Therefore, we use IoU to measure the positional relationship between objects in the association stage to avoid incorrect matching.

\bmhead{Association} Since TCBTrack no longer recalls low-confidence candidate boxes through a model, we adopt the BYTE method of ByteTrack\citep{zhang2022bytetrack}: instead of using a threshold to filter candidate boxes, we differentiate high-confidence and low-confidence candidate boxes using the threshold. Based on the observation that occlusion is often caused by two objects with different motion characteristics, we incorporate appearance score into the calculation of association weights. We define the weights as follows:
\begin{equation}
\label{20}
    Score=IoU*Det\_score*Temp\_score,
\end{equation}
\begin{equation}
\label{21}
    Temp\_score = Cosine(F_{1},F_{2}).
\end{equation}

\textit{IoU} is used to measure the motion characteristics between bounding boxes, while \textit{Det\_score} can treat boxes with low detection scores as unreliable objects during IoU matching to reduce the matching score. \textit{Temp\_score} is used to measure the appearance and temporal feature similarities between bounding boxes. We aim to set the matching weights in a way that allows for complementarity between motion characteristics, appearance features, and temporal features: When the IoU score is high, the challenge in MOT tasks lies in distinguishing the true object from surrounding objects, which can be addressed by appearance score. When the IoU score is low, we can also utilize it for recall. Moreover, when similar objects are connected, the temporal information in appearance feature can be used for differentiation. This strategy is based on the premise that the object features we extracted perform well in MOT tasks and the heatmap response is high. In section \ref{section4} we compare our strategy with linear weighting scores:
\begin{equation}
\label{13-}
    Score=(1-\delta)*IoU + \delta*Temp\_score
\end{equation}

Additionally, we use linear weighting method to update the features of the tracked object:
\begin{equation}
\label{22}
    F^t = (1-\gamma)*F^{t-1}+\gamma*F^t.
\end{equation}

\begin{algorithm}[!t]
\caption{Pseudocode of the tracking process under TCBTrack architecture.}
\begin{algorithmic}
    \State \hspace{0.2cm}{\textbf{Input}}: image frame $I^t$, track indices $\mathcal{T}^{t-1}=\{ \mathcal{T}^{t-1}_1, \mathcal{T}^{t-1}_2, ..., \mathcal{T}^{t-1}_n  \}$, model backbone $\Phi$, detection head $\Psi$, ReID head $\Omega$
\end{algorithmic}
\begin{algorithmic}[1]
    \State  Extract features $ f^t= \Phi(I^t)$
    \State $\mathcal{D}^t = \Psi(f^{t}),F^t = \Omega(f^t)$
    \State $[\mathcal{D}^t_{high},\mathcal{D}^t_{low}] \gets \mathcal{D}^t, [F^t_{high},F^t_{low}] \gets F^t$
    \Statex
    \State $\mathcal{T}^{t} \gets Association1(\mathcal{T}^{t-1},\mathcal{D}^t_{high},F^t)$
    
    \State $[\mathcal{D}^t_{remain}, \mathcal{T}^{t-1}_{remain}] \gets [\mathcal{T}^{t}, \mathcal{T}^{t-1}, \mathcal{D}^{t}_{high}]$
    \Statex
    \State $\mathcal{T}^{t} \gets Association2(\mathcal{T}^{t-1}_{remain},\mathcal{D}^t_{remain}) \bigcup \mathcal{T}^{t}$
     
     \State $[\mathcal{T}^{t-1}_{remain},\mathcal{D}^t_{remain}] \gets [\mathcal{T}^{t},\mathcal{T}^{t-1}_{remain},\mathcal{D}^t_{remain}]$
    \Statex
     \State $\mathcal{T}^{t} \gets Association3(\mathcal{T}^{t-1}_{remain},\mathcal{D}^t_{low}) \bigcup \mathcal{T}^{t}$
     
     \State $[\mathcal{D}^t_{remain2}, \mathcal{T}^{t-1}_{remain2}] \gets [\mathcal{T}^{t}, \mathcal{T}^{t-1}, \mathcal{D}^{t}_{low}]$
     \Statex
    \State $[\mathcal{T}^{t-1}_{lost},\mathcal{T}^{t-1}_{remove}] \gets \mathcal{T}^{t-1}_{remain2}$
    \State  $\mathcal{T}^{t-1}_{new} \gets \mathcal{D}^{t}_{remain}$
    \State $\mathcal{T}^{t} \gets \mathcal{T}^{t-1}_{lost}\bigcup \mathcal{T}^{t-1}_{new}\bigcup \mathcal{T}^{t}$
    
    \State\Return $\mathcal{T}^{t}$
\end{algorithmic}

\label{ag4}
\end{algorithm}
Here we set $\gamma$ to 0.1  because experiments show that small values of $\gamma$ reduce the incorrect matching of adjacent. However,  if $\gamma$ is too small, then the features of objects in the current frame cannot be updated. Also based on our training method, we use the cosine distance as the weight for the first stage matching. The entire tracking process is described in Algorithm \ref{ag4}: Three associations are used to match the current frame with the tracked object from the previous frame. \textit{Association1} uses equation \eqref{20} to match most of the objects, and then \textit{Association2} and \textit{Association3} use IoU metric to match the remaining objects from high confidence to low confidence. Unmatched high confidence boxes are added to the tracking sequence as new tracking objects. More detailed discussion of the hyper-parameters and the results on the data sets can be found in section \ref{section4}.

\begin{table*}[t]
\centering
\caption{Analysis of different training methods on MOT17 half-val set and DanceTrack validation set. We also compare different ways of using feature distance metrics. When using linear weighting, we use grid search to search for the best weights and display the best results in the table. }
\label{table1}
\begin{tabular}{|c|c|c|c|c|c|c|c|c|c|c|}
\hline
\multicolumn{2}{|c|}{\multirow{2}{*}{Training}} &\multirow{2}{*}{Association}& \multicolumn{3}{c|}{MOT17}&\multicolumn{3}{c|}{DanceTrack}\\
\cline{4-9}

\multicolumn{2}{|c|}{} & &  MOTA$\uparrow$ & HOTA$\uparrow$ & IDF1$\uparrow$ & MOTA$\uparrow$ & HOTA$\uparrow$ & IDF1$\uparrow$\\
\hline
\multicolumn{2}{|c|}{\multirow{3}{*}{ID Loss}} & Motion & 77.8 & 68.0 &
80.0 & 86.3 &	46.1&50.9    \\
\cline{3-9}

\multicolumn{2}{|c|}{} & +Linear &	77.6&67.8 &80.0	& 87.3	& 46.1 & 51.0			\\
\cline{4-9}
\multicolumn{2}{|c|}{} & +Product & 77.6 & 67.6 & 79.7 & 86.0 &	45.8 & 51.0  \\
\hline
\multicolumn{2}{|c|}{\multirow{3}{*}{Ours}} & Motion & 77.8 & 68.0 &
80.0 & 86.3 &	46.1&50.9    \\
\cline{3-9}

\multicolumn{2}{|c|}{} & +Linear &77.6	&67.6	&79.4	&86.4	&45.7&51.1			\\
\cline{4-9}
\multicolumn{2}{|c|}{} & +Product & 77.6 & 67.7 &
79.7 & 88.1 &	47.9 & 53.7  \\
\hline
\end{tabular}
\label{tab:loss}
\end{table*}

\section{Experiments}\label{section4}
\subsection{Implementation Details and Baseline Setup}
\subsubsection{Implementation of TCBTrack}
We choose YOLOX-X as the detector for TCBTrack because it has an acceptable model size and because it is widely used. This simplifies the comparisons with other methods. To investigate the performance of the model under different detectors, we also utilize YOLO-S and YOLO-M as detectors to observe the tracking performance as detector capabilities are gradually decreased. We use the pretrained models released by ByteTrack/DanceTrack and only train the embedding head. For the models of MOT17 that are not public, we train on the combination of MOT17 half-training set and CrowdHuman dataset\citep{shao2018crowdhuman}. Building upon the experiments conducted with OMC, we introduce the DanceTrack dataset to examine the model's robustness. This addition is motivated by the fact that the MOT16/17/20 datasets predominantly consist of surveillance videos in which objects mostly move in a simple, linear manner. DanceTrack, on the other hand, provides a more complex dataset with objects exhibiting dramatic variations, irregular motions, and higher levels of similarity among objects, as it was collected from dance videos. In our experiments, we primarily explored the performance of cross-correlation learning in terms of robustness and accuracy, using the: 
\begin{itemize}
    \item Compare different training methods to demonstrate the advantages of cross-correlation learning in MOT tasks.
    \item Show that our enhanced association approach outperforms methods that primarily rely on IoU.
    \item Conduct comparisons with other trackers under high-speed conditions by sampling frames. We utilize the MOT17 dataset and DanceTrack dataset, which have frame rates of 25 fps and 20 fps respectively. 
\end{itemize}

For MOT dataset, we divide the training dataset into two halves, using the first half(half-train) for training and the second half(half-val) for validation(See \cite{zhou2020tracking}). Consequently, We choose to compare our tracker with other trackers that rely on motion cues to show the improvement to the MOT task brought by the generated temporal features.

For the training of the detection models, we employ the training parameters from YOLOX. For training embedding head, separate training is conducted using only the training sets from each dataset. The training involves 20 epochs with a batch size of 16. The initial learning rate was set to 1e-3, and the optimizer used was SGD with a weight decay of 5e-4 and a momentum of 0.9. The entire model is trained on four NVIDIA Tesla A100 GPUs, and all of the experiments are conducted on a single NVIDIA Tesla A100 GPU and Xeon(R) Gold 6146 3.20GHz CPU.

\subsubsection{Metrics}
We employ MOTA, FP, FN metrics from CLEAR\citep{bernardin2008evaluating} as our primary evaluation measures, supplemented by HOTA, AssA, DetA in \cite{luiten2021hota} and IDF1\citep{ristani2016performance}.By analysing FN and FP, we can compare the number of false negatives and false positives produced by the models under different hyper-parameters. MOTA, derived from metrics such as FP and FN, is used to assess the detection quality and accuracy of the tracker. HOTA employs a more complex approach and is considered to be a more comprehensive measure of tracking performance. DetA, as the accuracy of detection, is also used to evaluate the performance of the detector; IDF1, AssA are utilized to evaluate the tracker's association capabilities.

\subsection{Experimental analyses of TCBTrack}
\subsubsection{Ablation study}

We compare our training method with ID Loss. On the MOT17 dataset, using motion as the linear assignment weight already shows excellent results when the tracker incorporates Kalman filtering. When Kalman filtering can already address the issue, using features generated by the JDE tracker for association actually impairs tracking performance. However, as the dataset becomes more complex, our training method can provide more accurate association. Using equation \eqref{20}, we can address the issue of inaccurate IoU scores to a greater extent. The results are shown in Table \ref{tab:loss}.

 \begin{table}[!t]
  \centering
  \caption{Analysis of different association methods on DanceTrack validation set.}
  \vspace{-1.6ex}
    \begin{tabular}{l|@{}c@{} @{}c@{} @{}c@{} @{}c@{} @{}c@{}}
      \toprule
      Association                        ~&~ HOTA   ~  &~ DetA     ~&~ AssA     ~&~ MOTA     ~&~ IDF1     \\\midrule
      IoU                        & 44.7     & 52.6     & 25.3     & 87.3     & 36.8     \\
      SORT                    & 47.8     & 74.0     &  31.0     & 88.2     & 48.3     \\
      DeepSORT                     & 45.8     & 70.9     & 29.7     & 87.1     & 46.8     \\
      MOTDT                         & 39.2     & 68.8     & 22.5     & 84.3     & 39.6     \\
      BYTE                            & 47.1     & 70.5     & 31.5     & 88.2     & 51.9     \\
      OC-SORT                        & 52.1     & \textbf{79.8}     & 35.3     & 87.3     & 51.6     \\ 
      \midrule
      
      Ours                           & \textbf{54.6}     & 78.9     & \textbf{38.0}     & \textbf{89.9}     & \textbf{54.7} 
      \\\bottomrule
    \end{tabular}
  \label{tab:dancetrack_sota}
\end{table}

\begin{table}[!t]
        \caption{Analysis of the use of Kalman filter($K$). Our association method has a similar performance without any prediction of the objects. } 
        \fontsize{8pt}{4mm}\selectfont
            \begin{tabular}{@{}c@{} | @{}c@{}   |  @{}c@{} @{}c@{} @{}c@{} @{}c@{} @{}c@{}}
                \toprule
                \#NUM ~&~~ $K$ ~~& ~~MOTA$\uparrow$ ~~& ~~IDF1$\uparrow$~  ~ &  ~~FP$\downarrow$~~ &  ~~FN$\downarrow$~~ & ~~IDs$\downarrow$~~ 
                \\
                \midrule
1  &\checkmark  & 77.6 & 79.7 & 3436 & \textbf{9187} & \textbf{166} \\
2  &  & \textbf{77.8} & \textbf{79.4} & \textbf{3403} & 9263 & 174 \\
                \bottomrule
            \end{tabular}
        \vspace{-1em}
        \label{tab:kalman}
\end{table}

\begin{table*}[!t]
        \caption{Analysis of different backbones on MOT17 half-val set.} 
            \begin{tabular*}{\textwidth}{@{\extracolsep\fill}@{}c@{} | @{}c@{} @{}c@{}  @{}c@{}  @{}c@{} @{}c@{} @{}c@{} }
                \toprule
                ~~Detector~~&~~MOTA$\uparrow$~~&~~IDF1$\uparrow$~~&~~HOTA$\uparrow$~~&~~Params$\downarrow$~~&~~~GFLOPs$\downarrow$~~~&FPS$\uparrow$
                \\
                \toprule
                YOLOX-S & 71.1 & 73.6 & 63.0 & 9.9M & 90.07 & 58.5 \\
				~~YOLOX-M~~  & 74.5 & 76.2 & 65.4 & 27.4M & 239.77 & 50.6\\
				~~YOLO-X~~  & 77.6 & 79.3 & 67.5 & 104.7M & 881.25 & 27.7\\
                \bottomrule
            \end{tabular*}
        \label{tab:lightweight}
\end{table*}

\begin{table*}[!t]
\caption{Comparison with the state-of-the-art online MOT systems on DanceTrack benchmark. We assess a tracker's real-time performance by evaluating whether its inference speed surpasses the frame rate of the dataset(20 fps). We report the corresponding official metrics. ↑/↓ indicate that higher/lower is better, respectively. The best scores for each method is marked in \bf{bold}.}
\begin{center}

\begin{tabular*}{\textwidth}{@{\extracolsep\fill}l|ccccccc} \toprule
Method  & HOTA$\uparrow$  &DetA$\uparrow$ & AssA$\uparrow$&MOTA$\uparrow$ &IDF1$\uparrow$  \\ \hline
\textit{Non-real-time:}     \\
CenterTrack\citep{zhou2020tracking}~ & 41.8 & 78.1 & 22.6 & 86.8 & 35.7   \\
TransTrack\citep{sun2020transtrack}~  & 45.5 & 75.9 & 27.5 & 88.4 & 45.2   \\
GTR\citep{zhou2022global}~ & 48.0 & 72.5 & 31.9 & 84.7 & 50.3   \\
StrongSORT++\citep{du2023strongsort}~  & 55.6 & 80.7 & 38.6 & 91.1 &  55.2   \\
Deep OC-SORT\citep{maggiolino2023deep}~  & 61.3 & 82.2 & 45.8 & \bf{92.3} & 61.5   \\
MOTR\citep{zeng2022motr}~  & 54.2 & 73.5 & 40.2 & 79.7 & 51.5   \\
MOTRv2\citep{zhang2023motrv2}~ & \bf{73.4} & \bf{83.7} & \bf{64.4} & 92.1 & \bf{76.0}   \\
\midrule
\textit{Online $\&$ real-time:}     \\
FairMOT\citep{zhang2021fairmot}~  & 39.7 & 66.7 & 23.8 & 82.2 &  40.8   \\
TraDes\citep{wu2021track}~  & 43.3 & 74.5 & 25.4 & 86.2 & 41.2   \\
ByteTrack\citep{zhang2022bytetrack}~  & 47.7 & 71.0 & 32.1 & 89.6 & 53.9   \\
QDTrack\citep{pang2021quasi}~  & 54.2 & 80.1 & 36.8 & 87.7 &  50.4   \\
OC-SORT\citep{cao2023observation}~  & 55.1 & 80.3 & 38.3 & 92.0 &  54.6   \\
\bf{TCBTrack(Ours)}~  & \bf{56.8}  & \bf{81.9} & \bf{39.5} & \bf{92.5}&\bf{58.1}   \\

\midrule
\textit{Offline:}     \\
SUSHI\citep{cetintas2023unifying}~  & \bf{63.3} & \bf{80.1} & \bf{50.1} & \bf{88.7} & \bf{63.4}   \\
\bottomrule
\end{tabular*}
\end{center}
\label{tab:comparison4}%
\end{table*}

We investigate the effectiveness of our method for association. The results are shown in Table \ref{tab:dancetrack_sota}. Almost all Tracking-by-Detection paradigms currently rely on IoU scores of bounding boxes between consecutive frames as the primary weighting factor for the association algorithm. As mentioned in \cite{zhang2021fairmot}, this is because factors such as size variations, lighting changes, and resolution in multi-object tracking make it challenging to distinguish object features. Methods such as BYTE perform well on simple MOT datasets, they fail when there are large errors in estimates of the movements of objects. We provide a more robust methods for obtaining distance measure, furtherly improving the tracking performance.

To investigate the removal of Kalman filtering from our method, a comparative experiment is conducted on the MOT17 half-val set(The MOT17 contains pedestrians whose movements which can be estimated accurately using Kalman filtering), shown in Table \ref{tab:kalman}. Our method compensates for the loss of IoU information, especially when the movements of the objects are large.

\begin{table*}[!t]
\caption{Comparison with the state-of-the-art online MOT systems of different types under private detection on MOT17 benchmark. We assess a tracker's real-time performance by evaluating whether its inference speed surpasses the average frame rate of the dataset(25 fps). We report the corresponding official metrics. ↑ indicates that higher is better, ↓ indicates that lower is better. The best scores for each method is marked in \bf{bold}.}
\begin{center}
\begin{tabular*}{\textwidth}{@{\extracolsep\fill}l|ccccccc} \toprule
Method  & HOTA$\uparrow$  &MOTA$\uparrow$ & IDF1$\uparrow$&IDs$\downarrow$ &FPS$\uparrow$  \\ \hline
\textit{Non-real-time:}     \\
Tracktor++\citep{bergmann2019tracking}~  & 44.8 & 56.3 & 55.1 & 1987 & 1.5   \\
CTracker\citep{peng2020chained}~  & 49.0 & 66.6 & 57.4 & 5529 & 6.8   \\
TubeTK\citep{pang2020tubetk}~  & 48.0 & 63.0 & 58.6 & 4137 &  3.0   \\
UMA\citep{yin2020unified}~  & - & 53.1 & 54.4 & 2251 & 5.0   \\
CenterTrack\citep{zhou2020tracking}~  & 52.2 & 67.8 & 64.7 & 3039 & 17.5   \\
SOTMOT\citep{zheng2021improving}~  & - & 71.0 & 71.9 & 5184 & 16.0   \\
TraDes\citep{wu2021track}~  & 52.7 & 69.1 & 63.9 & 3555 & 17.5   \\
SiamMOT\citep{shuai2021siammot}~  & - & 76.3 & 72.3 & - & 12.8   \\
QDTrack\citep{pang2021quasi}~  & 53.9 & 68.7 & 66.3 & 3378 & \bf{20.3}   \\
CorrTracker\citep{wang2021multiple}~  & 60.7 & 76.5 & 73.6 & 3369 & 15.6   \\
CSTrack\citep{liang2022rethinking}~  & 59.3 & 74.9 & 72.6 & 3567 & 15.8   \\
TransMOT\citep{chu2023transmot}~  & - & 76.7 & 75.1 & 2346 & 9.6   \\
MOTR\citep{zeng2022motr}~  & 57.8 & 73.4 & 68.6 & 2439 & 9.5   \\
MOTRv2\citep{zhang2023motrv2}~  & 62.0 & 78.6 & 75.0 & - & 6.9   \\
BoT-SORT\citep{aharon2022bot}~  & 64.6 & 80.6 & 79.5 & 1257 & 6.6   \\
UTM\citep{you2023utm}~  & 64.0 & \bf{81.8} & 78.7 & 1436 & 13.1   \\
StrongSORT\citep{du2023strongsort}~  & 63.5 & 78.3 & 78.5 & 1446 & 7.5   \\
Deep OC-SORT\citep{maggiolino2023deep}~  & \bf{64.9} & 79.4 & \bf{80.6} & \bf{1023} & \textless11.0   \\

\midrule
\textit{Online $\&$ real-time:}     \\
SORT\citep{bewley2016simple}~  & 34.0 & 43.1 & 39.8 & 4852 & \bf{143.3}   \\
FairMOT\citep{zhang2021fairmot}~ & 59.3 & 73.7 & 72.3 & 3303 & 25.9   \\
ByteTrack\citep{zhang2022bytetrack}~  & 63.1 & \bf{80.3} & 77.3 & 2196 & 29.6   \\
OC-SORT\citep{cao2023observation}~ & \bf{63.2} & 78.0 & \bf{77.5} & \bf{1950} & 29.0   \\
\bf{TCBTrack(Ours)}     & 62.1 & 79.3 & 75.8 & 2157 & 27.7    \\
\midrule
\textit{Offline:}     \\
MPNTrack\citep{braso2020learning}~  & 49.0 & 58.8 & 61.7 & 1185 & 6.5   \\
ReMOT\citep{yang2021remot}~  & 59.7 & 77.0 & 72.0 & 2853 & 1.8   \\
MAATrack\citep{stadler2022modelling}~  & 62.0 & 79.4 & 75.9 & 1452 & \bf{189.1}   \\
SUSHI\citep{cetintas2023unifying}~  & \bf{66.5} & \bf{81.1} & \bf{83.1} & \bf{1149} & 14.2   \\
\bottomrule
\end{tabular*}
\end{center}
\label{tab:comparison5}%
\end{table*}

\begin{table*}[!t]
\caption{Comparison with the state-of-the-art online MOT systems using different paradigms under private detection on MOT20 benchmark. We report the corresponding official metrics. ↑ indicates that higher is better, ↓ indicates that lower is better. The best scores of methods are marked in \bf{bold}.}
\begin{center}
\begin{tabular*}{\textwidth}{@{\extracolsep\fill}l|ccccccc} \toprule
Method  & HOTA$\uparrow$  &MOTA$\uparrow$ & IDF1$\uparrow$&IDs$\uparrow$ &FPS$\uparrow$  \\ \hline
\textit{JDE paradigm:}     \\
FairMOT\citep{zhang2021fairmot}~  & 54.6 & 61.8 & 67.3 & 5243 & 13.2   \\
CSTrack\citep{liang2022rethinking}~  & 54.0 & 66.6 & 68.6 & 3196 & 4.5   \\
GSDT\citep{wang2021joint}~  & 53.6 & 67.1 & 67.5 & 3131 & 0.9   \\
CorrTracker\citep{wang2021multiple}~  & - & 65.2 & 69.1 & 5183 & 8.5   \\
RelationTrack\citep{yu2022relationtrack}~  & 56.5 & 67.2 & 70.5 & 4243 & 2.7   \\
\bf{TCBTrack(ours)}~  & \bf{60.6} & \bf{76.0} & \bf{74.4} & \bf{1174} & \bf{16.1}   \\
\midrule
\textit{SDE paradigm:}     \\
BoT-SORT-ReID\citep{aharon2022bot}~  & 63.3 & \bf{77.8} & 77.5 & 1257 & 2.4   \\
StrongSORT\citep{du2023strongsort}~  & 61.5 & 72.2 & 75.9 & 1066 & 1.5   \\
Deep OC-SORT\citep{maggiolino2023deep}~  & \bf{63.9} & 75.6 & \bf{79.2} & \bf{779} & \textless5.0   \\

\midrule
\textit{Motion:}     \\
SORT\citep{bewley2016simple}~  & 36.1 & 42.7 & 45.1 & 4470 & \bf{57.3}   \\
ByteTrack\citep{zhang2022bytetrack}~  & 61.3 & 77.8 & 75.2 & 1223 & 17.5   \\
OC-SORT\citep{cao2023observation}~  & 62.1 & 75.5 & 75.9 & \bf{913} & -   \\
MotionTrack\citep{qin2023motiontrack}~  &\bf{62.8} & \bf{78.0} & 76.5 & 1165 & -   \\
UCMCTrack\citep{yi2023ucmctrack}~ & \bf{62.8} & 75.6 & \bf{77.4} & 1335 & -    \\
\midrule
\textit{Transformer paradigm:}     \\
MOTRv2\citep{zhang2023motrv2}~  & \bf{60.3} & \bf{76.2} & \bf{72.2} & - & \bf{6.9}   \\
\midrule
\textit{Offline:}     \\
MPNTrack\citep{braso2020learning}~  & 46.8 & 57.6 & 59.1 & 1210 & 6.5   \\
ReMOT\citep{yang2021remot}~  & 61.2 & \bf{77.4} & 73.1 & 1789 & 0.4   \\
MAATrack\citep{stadler2022modelling}~  & 57.3 & 73.9 & 71.2 & 1331 & \bf{14.7}   \\
SUSHI\citep{cetintas2023unifying}~  & \bf{64.3} & 74.3 & \bf{79.8} & \bf{706} & \textless10.0   \\
\midrule
\textit{Others:}     \\
SOTMOT\citep{zheng2021improving}~  & - & 68.6 & 71.4 & 4209 & \bf{8.5}   \\
UTM\citep{you2023utm}~ & \bf{62.5} & \bf{78.2} & \bf{76.9} & \bf{1228}&6.2    \\
\bottomrule
\end{tabular*}
\end{center}
\label{tab:comparison6}%
\end{table*}

\begin{figure}[!t]
\centering
\includegraphics[width=2.99in]{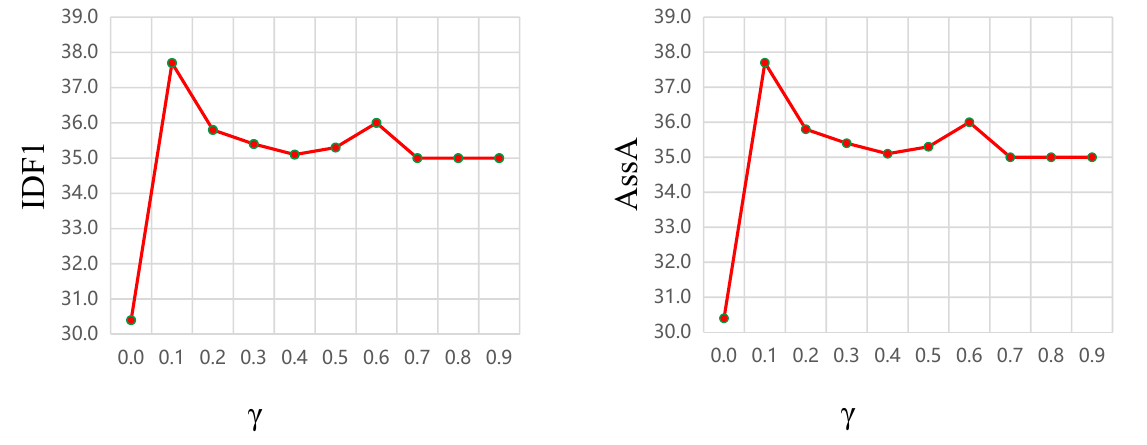}
\caption{Analysis of hyper-parameter $\gamma$ mentioned in equation (\ref{22}). Experiments are conducted on DanceTrack validation set. }
\label{fig_update}
\end{figure}

We also analyse the feature update strategy, with relevant results shown in Fig. \ref{fig_update}. A smaller update weight improves association performance, as identity switch often happens when the object are close to other objects. A larger update weight introduces significant noise from similar background information, which is not suitable for our association method. 

Furthermore, we investigate the feasibility of model lightweight. Table \ref{tab:lightweight} presents the performance of TCBTrack on MOT17 half-val set at different scales. It is observed that even when the detector becomes less accurate, our method still maintains tracking performance. For training sets such as MOT16, MOT17, DanceTrack, etc., we can achieve real-time requirements by lightweighting the detector. However, for the MOT20 dataset, we can only achieve online real-time performance by subsampling frames. This is because the MOT20 dataset contains a large number of objects in a single frame(up to around 200), causing tracking to consume a significant amount of time during the matching stage. This issue is also faced by all Tracking-by-Detection trackers. Possible solutions are: (1) Replace association with lightweight deep learning models. (2) Design a faster matching algorithm. This issue requires further exploration in future work. Currently, our model can address online and real-time tracking needs in most scenarios by model lightweight if necessary.

\subsubsection{Comparison with others}
\bmhead{DanceTrack dataset}We compare TCBTrack with existing tracking methods on the DanceTrack test set. Table \ref{tab:comparison4} shows that in the real-time tracker category, our method achieves state-of-the-art(SOTA) performance in metrics such as MOTA(+1.5), IDF1(+3.5), AssA(+1.2), and HOTA(+1.7). There is still some room for improvement to reach the current SOTA in non-real-time trackers.This is reasonable, because the trade-off between speed and accuracy inherently limits the achievable precision. In the existing SOTA, the most notable gap is observed between MOTRv2 and other methods. Our analysis suggests that this discrepancy arises from the fact that all trackers use IoU as the positional information measure for candidate bounding boxes during the tracking phase. However, in practical scenarios, IoU can significantly affect object matching due to inaccurate outputs from the detector. A single erroneous match may cause failure in the tracking. MOTRv2 addresses this issue by leveraging the transformer network to obtain robust positional information for the objects. In future research, we will focus on enhancing the association stage of Tracking-by-Detection paradigm trackers, aiming to achieve a true fusion of detection and tracking by improving the weighting scheme.

\bmhead{MOT dataset}We further validate the robustness of TCBTrack on different datasets by experimenting on MOT17(See Table \ref{tab:comparison5}) and MOT20(See Table \ref{tab:comparison6}). The tracking performance is on par with the best trackers across the board, which demonstrates the robustness of our approach. We have a slight performance gap (-2.8 HOTA, -0.1 MOTA, -4.8 IDF1) in MOT17 compared to Deep OC-SORT, a tracker that directly embeds the ReID model. However, we possess two distinct advantages over it: (1) Deep OC-SORT fails to achieve a balance between accuracy and speed due to its reliance on both high-performing detectors and large ReID models. (2) We are able to avoid incorporating ID loss during tracker training, which enables us to train on larger datasets with the same training epochs. When compared to other methods that rely on motion, although our algorithm does not achieve significant improvement on these two datasets, we are still the first tracker to fully utilize temporal cues for matching in the association stage and achieve good performance. In addition, we have comprehensively analysed the experimental results for the MOT dataset and the DanceTrack dataset. Our TCBTrack is more robust, accurate, and lightweight than other current trackers in various scenarios. We believe that temporal cues of the objects is still worth further investigation to achieve a better integration of ReID and MOT tasks.

\begin{table}[!t]
\caption{Comparison with the state-of-the-art online MOT systems on MOT17 ``high-speed'' dataset and DanceTrack ``high-speed'' dataset. We use ratios to show the fraction of the frames that are used. We report the corresponding official metrics. ↑ indicates that higher is better, ↓ indicates that lower is better. The best scores  and the least decreasing scores of methods are marked in \bf{bold}.}
\begin{tabular}{@{}c@{}|@{}c@{}|@{}c@{} @{}c@{} @{}c@{}} \toprule
Method ~&~ Ratio  ~&~ MOTA$\uparrow$  ~&~  IDF1$\uparrow$    ~&~  HOTA$\uparrow$ \\ \hline
\midrule  
\multicolumn{1}{c}{\bf{MOT17}}\\
\midrule  
SORT*\footnotemark[1]&1.0&74.7&77.7&66.4 \\
\citep{bewley2016simple}&0.5&71.2&76.4&64.9 \\
&0.33&68.1(-6.6)&72.7(-5.0)&62.1(-4.3) \\
\midrule
ByteTrack
&1.0&77.8&80.0&68.0\\
\citep{zhang2022bytetrack}&0.5&75.6&77.2&65.6\\
&0.33&\bf{73.9(-3.9)}&74.2(-5.8)&\textbf{63.7}(-5.3)\\
\midrule
OCSORT
&1.0&74.5&77.8&66.3\\
\citep{cao2023observation}&0.5&71.4&76.6&65.1\\
&0.33&68.7(-5.8)&74.1\bf{(-3.7)}&63.3\bf{(-3.0)}\\
\midrule
TCBTrack(ours)
&1.0&77.6&79.3&67.5\\
&0.5&75.6&76.7&65.4\\
&0.33&73.7\bf{(-3.9)}&\textbf{74.3}(-5.0)&63.4(-4.1)\\
\midrule  
\multicolumn{1}{c}{\bf{DanceTrack}}\\
\midrule  
SORT*\footnotemark[1]&1.0&85.0&48.8&47.3 \\
\citep{bewley2016simple}&0.5&81.8&44.3&41.9 \\
&0.33&76.8(-8.2)&39.4(-9.4)&37.0(-10.3) \\
\midrule
ByteTrack&1.0&86.3&46.1&50.9 \\
\citep{zhang2022bytetrack}&0.5&83.7&41.8&48.6 \\
&0.33&80.3(-6.0)&39.1(-7.0)&44.8(-6.1) \\
\midrule
OC-SORT&1.0&87.3&52.1&51.6 \\
\citep{cao2023observation}&0.5&84.6&48.2&47.9 \\
&0.33&80.8(-6.5)&44.2(-7.9)&42.8(-8.8) \\
\midrule
TCBTrack(ours)&1.0&89.9&54.7&54.6 \\
&0.5&88.8&52.1&53.4 \\
&0.33&\bf{87.1(-2.8)}&\bf{48.8(-5.8)}&\bf{49.5(-5.1)} \\
\bottomrule
\end{tabular}
\label{tab:comparison3}\footnotetext[1]{SORT*: We have replaced the detection head for SORT with YOLOX for fair comparison with other  Tracking-by-Detection trackers.}
\end{table}

\begin{figure}[!t]
\centering
\includegraphics[width=2.99in]{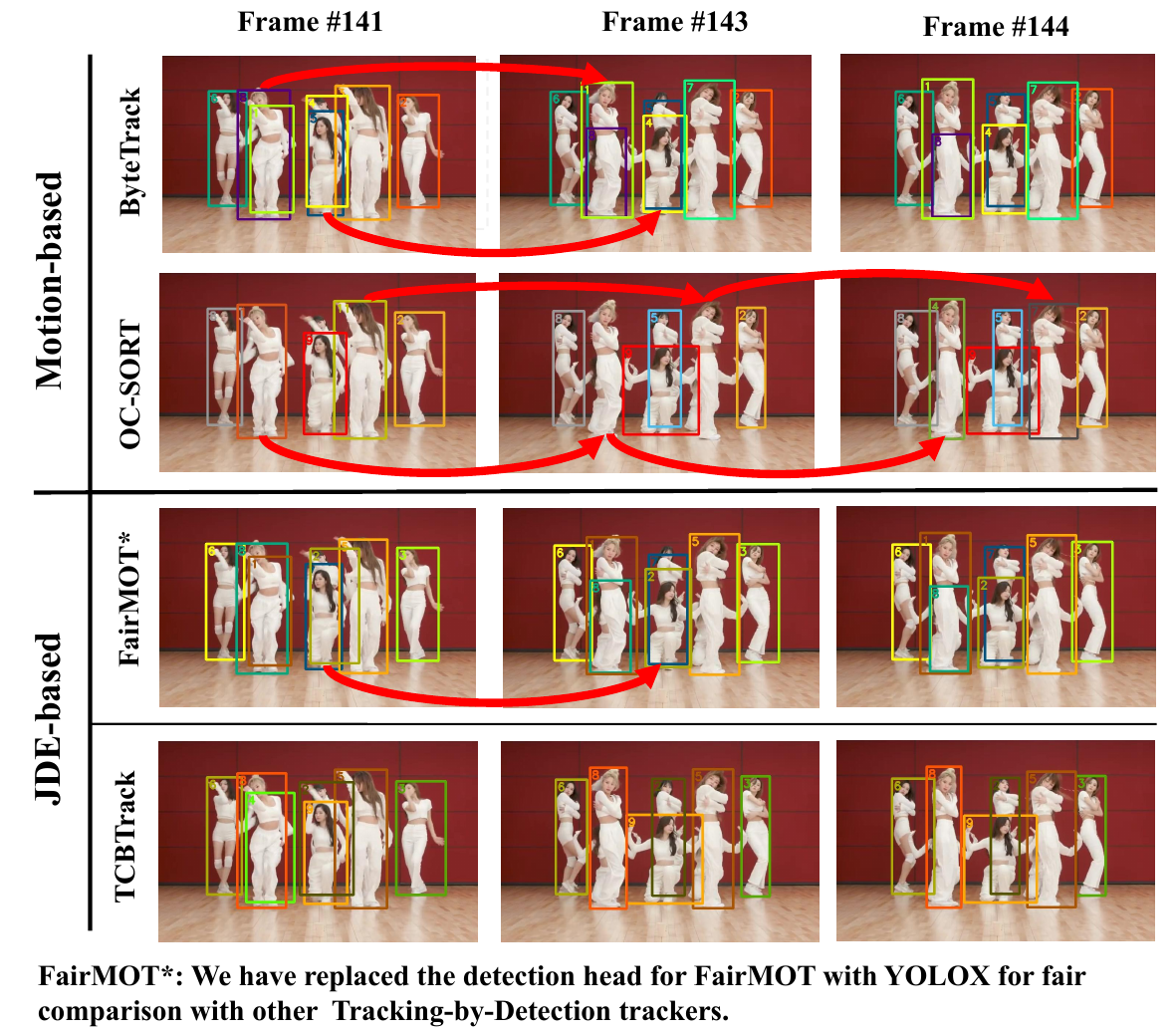}
\caption{Analysis of TCBTrack compared with FairMOT and OC-SORT on our modified DanceTrack dataset(0.33 ratio). The arrows indicate errors in the tracking process, such as identity switching or identity loss. }
\label{fig_comvis} 
\end{figure}

\begin{figure*}[!t]
\centering
\includegraphics[width=6.2in]{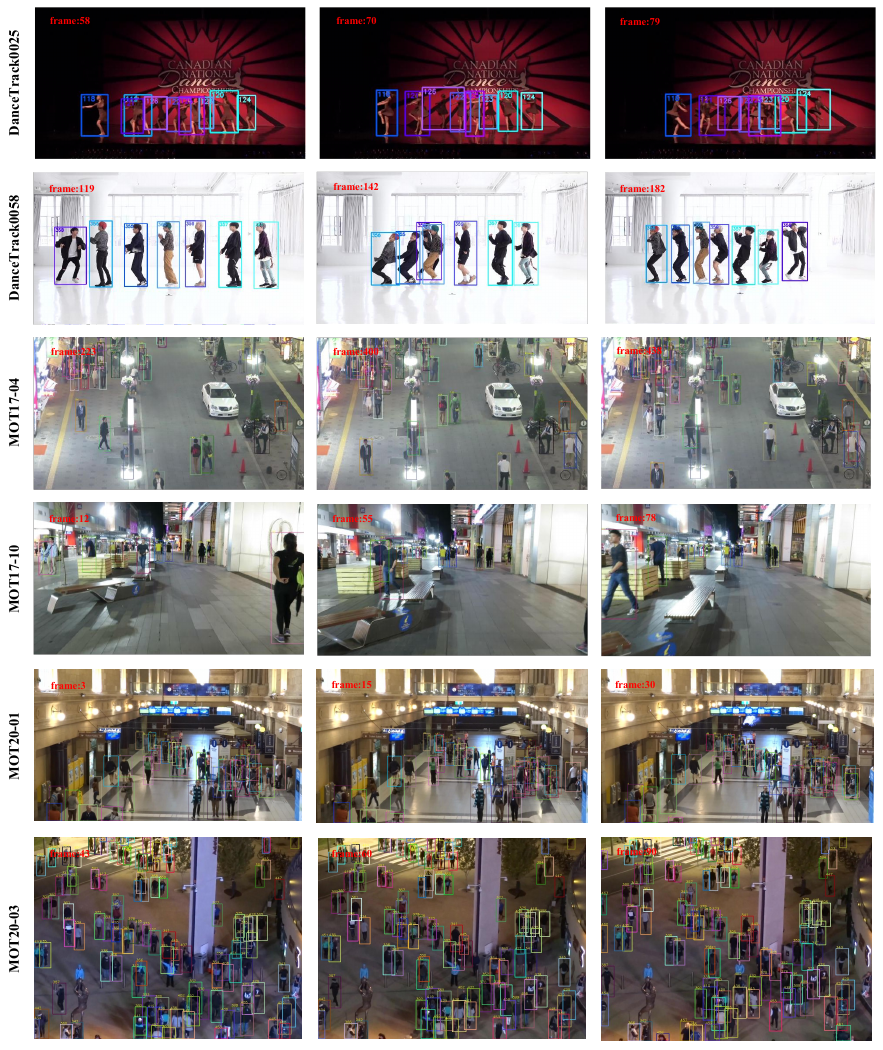}
\caption{ Sample tracking results of TCBTrack on DanceTrack validation set, MOT17 validation set and MOT20 validation set. Each bounding box is assigned to a dedicated ID and color for differentiation.}
\label{fig_vis}
\end{figure*}

\bmhead{Our modified dataset}Finally, we apply frame sampling to the MOT17 half-val dataset and DanceTrack validation set to construct a more difficult environment. As can be seen in Table \ref{tab:comparison3}, when evaluated on MOT17(0.33 ratio), our method achieves comparable performance(-0.2 MOTA, +0.1 IDF1, -0.3 HOTA) to those methods that rely on motion. Frame sampling has only a small effect. However, when evaluated on the DanceTrack dataset, the advantages of TCBTrack become apparent: we outperform other trackers under various tracking metrics(+7.3 MOTA, +4.6 IDF1, +4.7 HOTA) and the decrease in performance after frame sampling is relatively small. This demonstrates that TCBTrack handles objects moving irregularly at high speed. and with irregular patterns, proving the robustness of our association approach. Furthermore, as observed in both MOT17 and DanceTrack, TCBTrack performs well, even when 67\% of frames are dropped. If necessary, TCBTrack can reduce the frame rate to meet the tracking task requirements for different choices of hardware. 

We visualize the robustness analysis compared with other trackers, shown in Fig. \ref{fig_comvis}. ByteTrack and OC-SORT, while leveraging advanced Kalman filtering, encounter tracking failures in complex scenarios due to inaccuracies in prediction. FairMOT, which employs JDE paradigm, is unable to make correct associations based on apparence features when faced with overlapping similar objects. Our TCBTrack effectively addresses the problem of appearance similarity and maintains high tracking accuracy and stability in complex scenes.

\bmhead{Visualization}

Fig. \ref{fig_vis} presents some tracking results on the MOT17, MOT20 and DanceTrack datasets. From the figure, it can be observed that our method effectively mitigates the impact of imprecise detector outputs. For example, in the first row of tracking results, ID 126 experiences a detection deviation at frame 70 but continues to be correctly associated in subsequent frames. Our method also addresses occlusion issues, as illustrated in the second row of visualizations where the leftmost person passes behind six other persons with significant occlusion. Finally, on the MOT17 dataset and MOT20 dataset, we demonstrate that our tracker performs robustly in various scenarios with different perspectives and resolutions, maintaining accurate object tracking.

\section{Conclusion}
In this paper, we explore the importance of temporal cues in MOT tasks, arguing that incorporating temporal information enhances object feature association, leading to improved performance across various datasets. By incorporating the JDE paradigm and cross-correlation learning in the training process, TCBTrack enhances the quality of object features. Our experimental results on various datasets show effectiveness of TCBTrack. To simulate a more difficult tracking environment, we employ two strategies: detector lightweighting and dataset frame sampling. We then compare our approach, TCBTrack, with other state-of-the-art trackers to show its superiority in online and in real time performance. Through these comprehensive experiments, we demonstrate that our model is a robust and high-speed tracker, achieving top performance in online tracking scenarios. Overall, we believe that our work is a significant contribution to MOT that emphasises the role of temporal cues and presenting a versatile tracking system that strikes a balance between speed and accuracy.










\bibliography{sn-bibliography}

\begin{thebibliography}{93}
\providecommand{\natexlab}[1]{#1}
\providecommand{\url}[1]{{#1}}
\providecommand{\urlprefix}{URL }
\providecommand{\doi}[1]{\url{https://doi.org/#1}}
\providecommand{\eprint}[2][]{\url{#2}}
 \bibcommenthead

\bibitem[{Aharon et~al(2022)Aharon, Orfaig, and Bobrovsky}]{aharon2022bot}
Aharon N, Orfaig R, Bobrovsky BZ (2022) Bot-sort: Robust associations multi-pedestrian tracking. arXiv preprint arXiv:220614651

\bibitem[{Bergmann et~al(2019)Bergmann, Meinhardt, and Leal-Taixe}]{bergmann2019tracking}
Bergmann P, Meinhardt T, Leal-Taixe L (2019) Tracking without bells and whistles. In: Proceedings of the IEEE/CVF International Conference on Computer Vision, pp 941--951

\bibitem[{Bernardin and Stiefelhagen(2008)}]{bernardin2008evaluating}
Bernardin K, Stiefelhagen R (2008) Evaluating multiple object tracking performance: the clear mot metrics. EURASIP Journal on Image and Video Processing 2008:1--10

\bibitem[{Bertinetto et~al(2016)Bertinetto, Valmadre, Henriques, Vedaldi, and Torr}]{bertinetto2016fully}
Bertinetto L, Valmadre J, Henriques JF, et~al (2016) Fully-convolutional siamese networks for object tracking. In: Computer Vision--ECCV 2016 Workshops: Amsterdam, The Netherlands, October 8-10 and 15-16, 2016, Proceedings, Part II 14, Springer, pp 850--865

\bibitem[{Bewley et~al(2016)Bewley, Ge, Ott, Ramos, and Upcroft}]{bewley2016simple}
Bewley A, Ge Z, Ott L, et~al (2016) Simple online and realtime tracking. In: 2016 IEEE international conference on image processing (ICIP), IEEE, pp 3464--3468

\bibitem[{Bochinski et~al(2017)Bochinski, Eiselein, and Sikora}]{bochinski2017high}
Bochinski E, Eiselein V, Sikora T (2017) High-speed tracking-by-detection without using image information. In: 2017 14th IEEE international conference on advanced video and signal based surveillance (AVSS), IEEE, pp 1--6

\bibitem[{Bochinski et~al(2018)Bochinski, Senst, and Sikora}]{bochinski2018extending}
Bochinski E, Senst T, Sikora T (2018) Extending iou based multi-object tracking by visual information. In: 2018 15th IEEE International Conference on Advanced Video and Signal Based Surveillance (AVSS), IEEE, pp 1--6

\bibitem[{Bras{\'o} and Leal-Taix{\'e}(2020)}]{braso2020learning}
Bras{\'o} G, Leal-Taix{\'e} L (2020) Learning a neural solver for multiple object tracking. In: Proceedings of the IEEE/CVF conference on computer vision and pattern recognition, pp 6247--6257

\bibitem[{Cao et~al(2023)Cao, Pang, Weng, Khirodkar, and Kitani}]{cao2023observation}
Cao J, Pang J, Weng X, et~al (2023) Observation-centric sort: Rethinking sort for robust multi-object tracking. In: Proceedings of the IEEE/CVF Conference on Computer Vision and Pattern Recognition, pp 9686--9696

\bibitem[{Carion et~al(2020)Carion, Massa, Synnaeve, Usunier, Kirillov, and Zagoruyko}]{carion2020end}
Carion N, Massa F, Synnaeve G, et~al (2020) End-to-end object detection with transformers. In: European conference on computer vision, Springer, pp 213--229

\bibitem[{Cetintas et~al(2023)Cetintas, Bras{\'o}, and Leal-Taix{\'e}}]{cetintas2023unifying}
Cetintas O, Bras{\'o} G, Leal-Taix{\'e} L (2023) Unifying short and long-term tracking with graph hierarchies. In: Proceedings of the IEEE/CVF Conference on Computer Vision and Pattern Recognition, pp 22877--22887

\bibitem[{Chen et~al(2018)Chen, Ai, Zhuang, and Shang}]{chen2018real}
Chen L, Ai H, Zhuang Z, et~al (2018) Real-time multiple people tracking with deeply learned candidate selection and person re-identification. In: 2018 IEEE international conference on multimedia and expo (ICME), IEEE, pp 1--6

\bibitem[{Chu et~al(2023)Chu, Wang, You, Ling, and Liu}]{chu2023transmot}
Chu P, Wang J, You Q, et~al (2023) Transmot: Spatial-temporal graph transformer for multiple object tracking. In: Proceedings of the IEEE/CVF Winter Conference on Applications of Computer Vision, pp 4870--4880

\bibitem[{Chu et~al(2017)Chu, Ouyang, Li, Wang, Liu, and Yu}]{chu2017online}
Chu Q, Ouyang W, Li H, et~al (2017) Online multi-object tracking using cnn-based single object tracker with spatial-temporal attention mechanism. In: Proceedings of the IEEE international conference on computer vision, pp 4836--4845

\bibitem[{Cui et~al(2023)Cui, Zeng, Zhao, Yang, Wu, and Wang}]{cui2023sportsmot}
Cui Y, Zeng C, Zhao X, et~al (2023) Sportsmot: A large multi-object tracking dataset in multiple sports scenes. arXiv preprint arXiv:230405170

\bibitem[{Dai et~al(2021)Dai, Weng, Choi, Zhang, He, and Ding}]{dai2021learning}
Dai P, Weng R, Choi W, et~al (2021) Learning a proposal classifier for multiple object tracking. In: Proceedings of the IEEE/CVF Conference on Computer Vision and Pattern Recognition, pp 2443--2452

\bibitem[{Dendorfer et~al(2020)Dendorfer, Rezatofighi, Milan, Shi, Cremers, Reid, Roth, Schindler, and Leal-Taix{\'e}}]{dendorfer2020mot20}
Dendorfer P, Rezatofighi H, Milan A, et~al (2020) Mot20: A benchmark for multi object tracking in crowded scenes. arXiv preprint arXiv:200309003

\bibitem[{Dosovitskiy et~al(2020)Dosovitskiy, Beyer, Kolesnikov, Weissenborn, Zhai, Unterthiner, Dehghani, Minderer, Heigold, Gelly et~al}]{dosovitskiy2020image}
Dosovitskiy A, Beyer L, Kolesnikov A, et~al (2020) An image is worth 16x16 words: Transformers for image recognition at scale. arXiv preprint arXiv:201011929

\bibitem[{Du et~al(2023)Du, Zhao, Song, Zhao, Su, Gong, and Meng}]{du2023strongsort}
Du Y, Zhao Z, Song Y, et~al (2023) Strongsort: Make deepsort great again. IEEE Transactions on Multimedia

\bibitem[{Gao and Wang(2023)}]{gao2023memotr}
Gao R, Wang L (2023) Memotr: Long-term memory-augmented transformer for multi-object tracking. In: Proceedings of the IEEE/CVF International Conference on Computer Vision, pp 9901--9910

\bibitem[{Ge et~al(2021)Ge, Liu, Wang, Li, and Sun}]{ge2021yolox}
Ge Z, Liu S, Wang F, et~al (2021) Yolox: Exceeding yolo series in 2021. arXiv preprint arXiv:210708430

\bibitem[{Geiger et~al(2013)Geiger, Lenz, Stiller, and Urtasun}]{geiger2013vision}
Geiger A, Lenz P, Stiller C, et~al (2013) Vision meets robotics: The kitti dataset. The International Journal of Robotics Research 32(11):1231--1237

\bibitem[{Girshick(2015)}]{girshick2015fast}
Girshick R (2015) Fast r-cnn. In: Proceedings of the IEEE international conference on computer vision, pp 1440--1448

\bibitem[{Gu et~al(2019)Gu, Wang, and Hwang}]{gu2019efficient}
Gu R, Wang G, Hwang JN (2019) Efficient multi-person hierarchical 3d pose estimation for autonomous driving. In: 2019 IEEE Conference on Multimedia Information Processing and Retrieval (MIPR), IEEE, pp 163--168

\bibitem[{Hatamizadeh et~al(2022)Hatamizadeh, Tang, Nath, Yang, Myronenko, Landman, Roth, and Xu}]{hatamizadeh2022unetr}
Hatamizadeh A, Tang Y, Nath V, et~al (2022) Unetr: Transformers for 3d medical image segmentation. In: Proceedings of the IEEE/CVF winter conference on applications of computer vision, pp 574--584

\bibitem[{He et~al(2017)He, Gkioxari, Doll{\'a}r, and Girshick}]{he2017mask}
He K, Gkioxari G, Doll{\'a}r P, et~al (2017) Mask r-cnn. In: Proceedings of the IEEE international conference on computer vision, pp 2961--2969

\bibitem[{He et~al(2016)He, Takeuchi, Ninomiya, and Kato}]{he2016precise}
He M, Takeuchi E, Ninomiya Y, et~al (2016) Precise and efficient model-based vehicle tracking method using rao-blackwellized and scaling series particle filters. In: 2016 IEEE/RSJ international conference on intelligent robots and systems (IROS), IEEE, pp 117--124

\bibitem[{Kalman et~al(1960)}]{kalman1960contributions}
Kalman RE, et~al (1960) Contributions to the theory of optimal control. Bol soc mat mexicana 5(2):102--119

\bibitem[{Kim et~al(2022)Kim, Bras{\'o}, O{\v{s}}ep, and Leal-Taix{\'e}}]{kim2022polarmot}
Kim A, Bras{\'o} G, O{\v{s}}ep A, et~al (2022) Polarmot: How far can geometric relations take us in 3d multi-object tracking? In: European Conference on Computer Vision, Springer, pp 41--58

\bibitem[{Kuhn(1955)}]{kuhn1955hungarian}
Kuhn HW (1955) The hungarian method for the assignment problem. Naval research logistics quarterly 2(1-2):83--97

\bibitem[{Li et~al(2018)Li, Yan, Wu, Zhu, and Hu}]{li2018high}
Li B, Yan J, Wu W, et~al (2018) High performance visual tracking with siamese region proposal network. In: Proceedings of the IEEE conference on computer vision and pattern recognition, pp 8971--8980

\bibitem[{Li et~al(2019)Li, Wu, Wang, Zhang, Xing, and Yan}]{li2019siamrpn++}
Li B, Wu W, Wang Q, et~al (2019) Siamrpn++: Evolution of siamese visual tracking with very deep networks. In: Proceedings of the IEEE/CVF conference on computer vision and pattern recognition, pp 4282--4291

\bibitem[{Li et~al(2020)Li, Gao, and Jiang}]{li2020graph}
Li J, Gao X, Jiang T (2020) Graph networks for multiple object tracking. In: Proceedings of the IEEE/CVF winter conference on applications of computer vision, pp 719--728

\bibitem[{Li et~al(2023)Li, Yu, Philion, Anandkumar, Fidler, Jia, and Alvarez}]{li2023end}
Li Y, Yu Z, Philion J, et~al (2023) End-to-end 3d tracking with decoupled queries. In: Proceedings of the IEEE/CVF International Conference on Computer Vision, pp 18302--18311

\bibitem[{Liang et~al(2022{\natexlab{a}})Liang, Zhang, Zhou, Li, and Hu}]{liang2022one}
Liang C, Zhang Z, Zhou X, et~al (2022{\natexlab{a}}) One more check: making “fake background” be tracked again. In: Proceedings of the AAAI Conference on Artificial Intelligence, pp 1546--1554

\bibitem[{Liang et~al(2022{\natexlab{b}})Liang, Zhang, Zhou, Li, Zhu, and Hu}]{liang2022rethinking}
Liang C, Zhang Z, Zhou X, et~al (2022{\natexlab{b}}) Rethinking the competition between detection and reid in multiobject tracking. IEEE Transactions on Image Processing 31:3182--3196

\bibitem[{Lin et~al(2017{\natexlab{a}})Lin, Doll{\'a}r, Girshick, He, Hariharan, and Belongie}]{lin2017feature}
Lin TY, Doll{\'a}r P, Girshick R, et~al (2017{\natexlab{a}}) Feature pyramid networks for object detection. In: Proceedings of the IEEE conference on computer vision and pattern recognition, pp 2117--2125

\bibitem[{Lin et~al(2017{\natexlab{b}})Lin, Goyal, Girshick, He, and Doll{\'a}r}]{lin2017focal}
Lin TY, Goyal P, Girshick R, et~al (2017{\natexlab{b}}) Focal loss for dense object detection. In: Proceedings of the IEEE international conference on computer vision, pp 2980--2988

\bibitem[{Liu and Abbeel(2024)}]{liu2024blockwise}
Liu H, Abbeel P (2024) Blockwise parallel transformers for large context models. Advances in Neural Information Processing Systems 36

\bibitem[{Liu et~al(2023{\natexlab{a}})Liu, Zaharia, and Abbeel}]{liu2023ring}
Liu H, Zaharia M, Abbeel P (2023{\natexlab{a}}) Ring attention with blockwise transformers for near-infinite context. arXiv preprint arXiv:231001889

\bibitem[{Liu et~al(2020)Liu, Chu, Liu, and Yu}]{liu2020gsm}
Liu Q, Chu Q, Liu B, et~al (2020) Gsm: Graph similarity model for multi-object tracking. In: IJCAI, pp 530--536

\bibitem[{Liu et~al(2023{\natexlab{b}})Liu, Wang, Wang, Liu, and Bai}]{liu2023sparsetrack}
Liu Z, Wang X, Wang C, et~al (2023{\natexlab{b}}) Sparsetrack: Multi-object tracking by performing scene decomposition based on pseudo-depth. arXiv preprint arXiv:230605238

\bibitem[{Lu et~al(2020)Lu, Rathod, Votel, and Huang}]{lu2020retinatrack}
Lu Z, Rathod V, Votel R, et~al (2020) Retinatrack: Online single stage joint detection and tracking. In: Proceedings of the IEEE/CVF conference on computer vision and pattern recognition, pp 14668--14678

\bibitem[{Luiten et~al(2021)Luiten, Osep, Dendorfer, Torr, Geiger, Leal-Taix{\'e}, and Leibe}]{luiten2021hota}
Luiten J, Osep A, Dendorfer P, et~al (2021) Hota: A higher order metric for evaluating multi-object tracking. International journal of computer vision 129:548--578

\bibitem[{Luo et~al(2021)Luo, Yang, and Yuille}]{luo2021exploring}
Luo C, Yang X, Yuille A (2021) Exploring simple 3d multi-object tracking for autonomous driving. In: Proceedings of the IEEE/CVF International Conference on Computer Vision, pp 10488--10497

\bibitem[{Maggiolino et~al(2023)Maggiolino, Ahmad, Cao, and Kitani}]{maggiolino2023deep}
Maggiolino G, Ahmad A, Cao J, et~al (2023) Deep oc-sort: Multi-pedestrian tracking by adaptive re-identification. arXiv preprint arXiv:230211813

\bibitem[{Mahalanobis(2018)}]{mahalanobis2018generalized}
Mahalanobis PC (2018) On the generalized distance in statistics. Sankhy{\=a}: The Indian Journal of Statistics, Series A (2008-) 80:S1--S7

\bibitem[{Milan et~al(2016)Milan, Leal-Taix{\'e}, Reid, Roth, and Schindler}]{milan2016mot16}
Milan A, Leal-Taix{\'e} L, Reid I, et~al (2016) Mot16: A benchmark for multi-object tracking. arXiv preprint arXiv:160300831

\bibitem[{Pang et~al(2020)Pang, Li, Zhang, Li, and Lu}]{pang2020tubetk}
Pang B, Li Y, Zhang Y, et~al (2020) Tubetk: Adopting tubes to track multi-object in a one-step training model. In: Proceedings of the IEEE/CVF conference on computer vision and pattern recognition, pp 6308--6318

\bibitem[{Pang et~al(2021)Pang, Qiu, Li, Chen, Li, Darrell, and Yu}]{pang2021quasi}
Pang J, Qiu L, Li X, et~al (2021) Quasi-dense similarity learning for multiple object tracking. In: Proceedings of the IEEE/CVF conference on computer vision and pattern recognition, pp 164--173

\bibitem[{Papakis et~al(2020)Papakis, Sarkar, and Karpatne}]{papakis2020gcnnmatch}
Papakis I, Sarkar A, Karpatne A (2020) Gcnnmatch: Graph convolutional neural networks for multi-object tracking via sinkhorn normalization. arXiv preprint arXiv:201000067

\bibitem[{Peng et~al(2020)Peng, Wang, Wan, Wu, Wang, Tai, Wang, Li, Huang, and Fu}]{peng2020chained}
Peng J, Wang C, Wan F, et~al (2020) Chained-tracker: Chaining paired attentive regression results for end-to-end joint multiple-object detection and tracking. In: Computer Vision--ECCV 2020: 16th European Conference, Glasgow, UK, August 23--28, 2020, Proceedings, Part IV 16, Springer, pp 145--161

\bibitem[{Petrovskaya and Thrun(2009)}]{petrovskaya2009model}
Petrovskaya A, Thrun S (2009) Model based vehicle detection and tracking for autonomous urban driving. Autonomous Robots 26(2-3):123--139

\bibitem[{Qin et~al(2023)Qin, Zhou, Wang, Duan, Hua, and Tang}]{qin2023motiontrack}
Qin Z, Zhou S, Wang L, et~al (2023) Motiontrack: Learning robust short-term and long-term motions for multi-object tracking. In: Proceedings of the IEEE/CVF conference on computer vision and pattern recognition, pp 17939--17948

\bibitem[{Redmon and Farhadi(2018)}]{redmon2018yolov3}
Redmon J, Farhadi A (2018) Yolov3: An incremental improvement. arXiv preprint arXiv:180402767

\bibitem[{Ren et~al(2015)Ren, He, Girshick, and Sun}]{ren2015faster}
Ren S, He K, Girshick R, et~al (2015) Faster r-cnn: Towards real-time object detection with region proposal networks. Advances in neural information processing systems 28

\bibitem[{Ristani et~al(2016)Ristani, Solera, Zou, Cucchiara, and Tomasi}]{ristani2016performance}
Ristani E, Solera F, Zou R, et~al (2016) Performance measures and a data set for multi-target, multi-camera tracking. In: European conference on computer vision, Springer, pp 17--35

\bibitem[{Sadeghian et~al(2017)Sadeghian, Alahi, and Savarese}]{sadeghian2017tracking}
Sadeghian A, Alahi A, Savarese S (2017) Tracking the untrackable: Learning to track multiple cues with long-term dependencies. In: Proceedings of the IEEE international conference on computer vision, pp 300--311

\bibitem[{Schroff et~al(2015)Schroff, Kalenichenko, and Philbin}]{schroff2015facenet}
Schroff F, Kalenichenko D, Philbin J (2015) Facenet: A unified embedding for face recognition and clustering. In: Proceedings of the IEEE conference on computer vision and pattern recognition, pp 815--823

\bibitem[{Shao et~al(2018)Shao, Zhao, Li, Xiao, Yu, Zhang, and Sun}]{shao2018crowdhuman}
Shao S, Zhao Z, Li B, et~al (2018) Crowdhuman: A benchmark for detecting human in a crowd. arXiv preprint arXiv:180500123

\bibitem[{Shuai et~al(2021)Shuai, Berneshawi, Li, Modolo, and Tighe}]{shuai2021siammot}
Shuai B, Berneshawi A, Li X, et~al (2021) Siammot: Siamese multi-object tracking. In: Proceedings of the IEEE/CVF conference on computer vision and pattern recognition, pp 12372--12382

\bibitem[{Stadler and Beyerer(2022)}]{stadler2022modelling}
Stadler D, Beyerer J (2022) Modelling ambiguous assignments for multi-person tracking in crowds. In: Proceedings of the IEEE/CVF winter conference on applications of computer vision, pp 133--142

\bibitem[{Sun et~al(2020)Sun, Cao, Jiang, Zhang, Xie, Yuan, Wang, and Luo}]{sun2020transtrack}
Sun P, Cao J, Jiang Y, et~al (2020) Transtrack: Multiple object tracking with transformer. arXiv preprint arXiv:201215460

\bibitem[{Sun et~al(2022)Sun, Cao, Jiang, Yuan, Bai, Kitani, and Luo}]{sun2022dancetrack}
Sun P, Cao J, Jiang Y, et~al (2022) Dancetrack: Multi-object tracking in uniform appearance and diverse motion. In: Proceedings of the IEEE/CVF Conference on Computer Vision and Pattern Recognition, pp 20993--21002

\bibitem[{Tang et~al(2017)Tang, Andriluka, Andres, and Schiele}]{tang2017multiple}
Tang S, Andriluka M, Andres B, et~al (2017) Multiple people tracking by lifted multicut and person re-identification. In: Proceedings of the IEEE conference on computer vision and pattern recognition, pp 3539--3548

\bibitem[{Varior et~al(2016)Varior, Shuai, Lu, Xu, and Wang}]{varior2016siamese}
Varior RR, Shuai B, Lu J, et~al (2016) A siamese long short-term memory architecture for human re-identification. In: Computer Vision--ECCV 2016: 14th European Conference, Amsterdam, The Netherlands, October 11--14, 2016, Proceedings, Part VII 14, Springer, pp 135--153

\bibitem[{Wang et~al(2021{\natexlab{a}})Wang, Zheng, Pan, and Xu}]{wang2021multiple}
Wang Q, Zheng Y, Pan P, et~al (2021{\natexlab{a}}) Multiple object tracking with correlation learning. In: Proceedings of the IEEE/CVF Conference on Computer Vision and Pattern Recognition, pp 3876--3886

\bibitem[{Wang et~al(2021{\natexlab{b}})Wang, Kitani, and Weng}]{wang2021joint}
Wang Y, Kitani K, Weng X (2021{\natexlab{b}}) Joint object detection and multi-object tracking with graph neural networks. In: 2021 IEEE International Conference on Robotics and Automation (ICRA), IEEE, pp 13708--13715

\bibitem[{Wang(2022)}]{wang2022smiletrack}
Wang YH (2022) Smiletrack: Similarity learning for multiple object tracking. arXiv preprint arXiv:221108824

\bibitem[{Wang et~al(2020)Wang, Zheng, Liu, Li, and Wang}]{wang2020towards}
Wang Z, Zheng L, Liu Y, et~al (2020) Towards real-time multi-object tracking. In: European Conference on Computer Vision, Springer, pp 107--122

\bibitem[{Wojke et~al(2017)Wojke, Bewley, and Paulus}]{wojke2017simple}
Wojke N, Bewley A, Paulus D (2017) Simple online and realtime tracking with a deep association metric. In: 2017 IEEE international conference on image processing (ICIP), IEEE, pp 3645--3649

\bibitem[{Wu et~al(2021)Wu, Cao, Song, Wang, Yang, and Yuan}]{wu2021track}
Wu J, Cao J, Song L, et~al (2021) Track to detect and segment: An online multi-object tracker. In: Proceedings of the IEEE/CVF conference on computer vision and pattern recognition, pp 12352--12361

\bibitem[{Wu et~al(2019)Wu, Kong, Wang, Li, and Yin}]{wu2019unsupervised}
Wu Y, Kong D, Wang S, et~al (2019) An unsupervised real-time framework of human pose tracking from range image sequences. IEEE Transactions on Multimedia 22(8):2177--2190

\bibitem[{Yang et~al(2021)Yang, Chang, Sakti, Wu, and Nakamura}]{yang2021remot}
Yang F, Chang X, Sakti S, et~al (2021) Remot: A model-agnostic refinement for multiple object tracking. Image and Vision Computing 106:104091

\bibitem[{Yang et~al(2023)Yang, Han, Yan, Zhang, Qi, Lu, and Wang}]{yang2023hybrid}
Yang M, Han G, Yan B, et~al (2023) Hybrid-sort: Weak cues matter for online multi-object tracking. arXiv preprint arXiv:230800783

\bibitem[{Yi et~al(2023)Yi, Luo, Luo, Huang, Wu, Hu, and Hao}]{yi2023ucmctrack}
Yi K, Luo K, Luo X, et~al (2023) Ucmctrack: Multi-object tracking with uniform camera motion compensation. arXiv preprint arXiv:231208952

\bibitem[{Yin et~al(2020)Yin, Wang, Meng, Yang, and Shen}]{yin2020unified}
Yin J, Wang W, Meng Q, et~al (2020) A unified object motion and affinity model for online multi-object tracking. In: Proceedings of the IEEE/CVF Conference on Computer Vision and Pattern Recognition, pp 6768--6777

\bibitem[{You et~al(2023)You, Yao, Bao, and Xu}]{you2023utm}
You S, Yao H, Bao Bk, et~al (2023) Utm: A unified multiple object tracking model with identity-aware feature enhancement. In: Proceedings of the IEEE/CVF Conference on Computer Vision and Pattern Recognition, pp 21876--21886

\bibitem[{Yu et~al(2022)Yu, Li, Han, and Wang}]{yu2022relationtrack}
Yu E, Li Z, Han S, et~al (2022) Relationtrack: Relation-aware multiple object tracking with decoupled representation. IEEE Transactions on Multimedia

\bibitem[{Yu et~al(2023)Yu, Wang, Li, Zhang, Zhang, and Tao}]{yu2023motrv3}
Yu E, Wang T, Li Z, et~al (2023) Motrv3: Release-fetch supervision for end-to-end multi-object tracking. arXiv preprint arXiv:230514298

\bibitem[{Yu et~al(2016{\natexlab{a}})Yu, Li, Li, Liu, Shi, and Yan}]{yu2016poi}
Yu F, Li W, Li Q, et~al (2016{\natexlab{a}}) Poi: Multiple object tracking with high performance detection and appearance feature. In: Computer Vision--ECCV 2016 Workshops: Amsterdam, The Netherlands, October 8-10 and 15-16, 2016, Proceedings, Part II 14, Springer, pp 36--42

\bibitem[{Yu et~al(2020)Yu, Chen, Wang, Xian, Chen, Liu, Madhavan, and Darrell}]{yu2020bdd100k}
Yu F, Chen H, Wang X, et~al (2020) Bdd100k: A diverse driving dataset for heterogeneous multitask learning. In: Proceedings of the IEEE/CVF conference on computer vision and pattern recognition, pp 2636--2645

\bibitem[{Yu et~al(2016{\natexlab{b}})Yu, Jiang, Wang, Cao, and Huang}]{yu2016unitbox}
Yu J, Jiang Y, Wang Z, et~al (2016{\natexlab{b}}) Unitbox: An advanced object detection network. In: Proceedings of the 24th ACM international conference on Multimedia, pp 516--520

\bibitem[{Zeng et~al(2022)Zeng, Dong, Zhang, Wang, Zhang, and Wei}]{zeng2022motr}
Zeng F, Dong B, Zhang Y, et~al (2022) Motr: End-to-end multiple-object tracking with transformer. In: European Conference on Computer Vision, Springer, pp 659--675

\bibitem[{Zhai et~al(2022)Zhai, Kolesnikov, Houlsby, and Beyer}]{zhai2022scaling}
Zhai X, Kolesnikov A, Houlsby N, et~al (2022) Scaling vision transformers. In: Proceedings of the IEEE/CVF Conference on Computer Vision and Pattern Recognition, pp 12104--12113

\bibitem[{Zhang et~al(2019)Zhang, Yu, Jiao, Xing, El~Ghaoui, and Jordan}]{zhang2019theoretically}
Zhang H, Yu Y, Jiao J, et~al (2019) Theoretically principled trade-off between robustness and accuracy. In: International conference on machine learning, PMLR, pp 7472--7482

\bibitem[{Zhang et~al(2021)Zhang, Wang, Wang, Zeng, and Liu}]{zhang2021fairmot}
Zhang Y, Wang C, Wang X, et~al (2021) Fairmot: On the fairness of detection and re-identification in multiple object tracking. International Journal of Computer Vision 129:3069--3087

\bibitem[{Zhang et~al(2022)Zhang, Sun, Jiang, Yu, Weng, Yuan, Luo, Liu, and Wang}]{zhang2022bytetrack}
Zhang Y, Sun P, Jiang Y, et~al (2022) Bytetrack: Multi-object tracking by associating every detection box. In: European Conference on Computer Vision, Springer, pp 1--21

\bibitem[{Zhang et~al(2023)Zhang, Wang, and Zhang}]{zhang2023motrv2}
Zhang Y, Wang T, Zhang X (2023) Motrv2: Bootstrapping end-to-end multi-object tracking by pretrained object detectors. In: Proceedings of the IEEE/CVF Conference on Computer Vision and Pattern Recognition, pp 22056--22065

\bibitem[{Zheng et~al(2021)Zheng, Tang, Chen, Zhu, Wang, and Lu}]{zheng2021improving}
Zheng L, Tang M, Chen Y, et~al (2021) Improving multiple object tracking with single object tracking. In: Proceedings of the IEEE/CVF Conference on Computer Vision and Pattern Recognition, pp 2453--2462

\bibitem[{Zhou et~al(2020)Zhou, Koltun, and Kr{\"a}henb{\"u}hl}]{zhou2020tracking}
Zhou X, Koltun V, Kr{\"a}henb{\"u}hl P (2020) Tracking objects as points. In: European conference on computer vision, Springer, pp 474--490

\bibitem[{Zhou et~al(2022)Zhou, Yin, Koltun, and Kr{\"a}henb{\"u}hl}]{zhou2022global}
Zhou X, Yin T, Koltun V, et~al (2022) Global tracking transformers. In: Proceedings of the IEEE/CVF Conference on Computer Vision and Pattern Recognition, pp 8771--8780

\bibitem[{Zhu et~al(2018)Zhu, Yang, Liu, Kim, Zhang, and Yang}]{zhu2018online}
Zhu J, Yang H, Liu N, et~al (2018) Online multi-object tracking with dual matching attention networks. In: Proceedings of the European conference on computer vision (ECCV), pp 366--382

\end{thebibliography}

\end{document}